\documentclass{article}

\usepackage[preprint, nonatbib]{neurips_2020}

\usepackage[utf8]{inputenc} 
\usepackage[T1]{fontenc}    
\usepackage{hyperref}       
\usepackage{url}            
\usepackage{booktabs}       
\usepackage{amsfonts}       
\usepackage{nicefrac}       
\usepackage{microtype}      
\usepackage[numbers]{natbib}

\title{Variational Autoencoding of PDE Inverse Problems}

\author{%
  Daniel J. Tait
  \\
  University of Warwick\\
  The Alan Turing Institute\\
  \texttt{dtait@turing.ac.uk} \\
  \And
  Theodoros Damoulas \\
    University of Warwick\\
    The Alan Turing Institute\\
  \texttt{tdamoulas@turing.ac.uk} \\
}

\usepackage{wrapfig}
\usepackage[font=small]{caption, subcaption}
\usepackage{amsmath, amsfonts, amssymb}
\usepackage{empheq}

\newcommand{\bx}{\mathbf{x}}
\newcommand{\bu}{\mathbf{u}}
\newcommand{\bz}{\mathbf{z}}
\newcommand{\bff}{\mathbf{f}}
\newcommand{\by}{\mathbf{y}}

\newcommand{\bA}{\mathbf{A}}

\newcommand{\KL}{\operatorname{KL}}

\newcommand{\WeakForm}{
    \mathbf{L}[\mathbf{z}]\mu(\mathbf{x}, \mathbf{z}) - \mathbf{f} }

\newcommand{\Uniform}[1]{\mathrm{Uniform}(#1)}

\DeclareMathOperator*{\argmin}{arg\,min}

\newcommand{\Nnodes}{N_{\text{nodes}}}
\usepackage[export]{adjustbox}
\usepackage{titlesec}

\titleformat{\subsubsection}[runin]
  {\normalfont\normalsize\bfseries}{\thesubsubsection}{.5em}{}

\begin{document}

\maketitle

\begin{abstract}
    Specifying a governing physical model in the presence of missing physics and recovering its parameters are two intertwined and fundamental problems in science. Modern machine learning allows one to circumvent these, via emulators and surrogates, but in doing so disregards prior knowledge and physical laws that are especially important for small data regimes, interpretability, and decision making. In this work we fold the mechanistic model into a flexible data-driven surrogate to arrive at a physically structured decoder network. This provides accelerated inference for the Bayesian inverse problem, and can act as a drop-in regulariser that encodes a-priori physical information. We employ the variational form of the PDE problem and introduce stochastic local approximations as a form of model based data augmentation. We demonstrate both the accuracy and increased computational efficiency of the framework on real world settings and structured spatial processes.
\end{abstract}

\section{Introduction}
\begin{wrapfigure}{r}{0.45\linewidth}
    \centering
    \begin{subfigure}[b]{\linewidth}
    \centering
        \includegraphics[width=.5\textwidth]{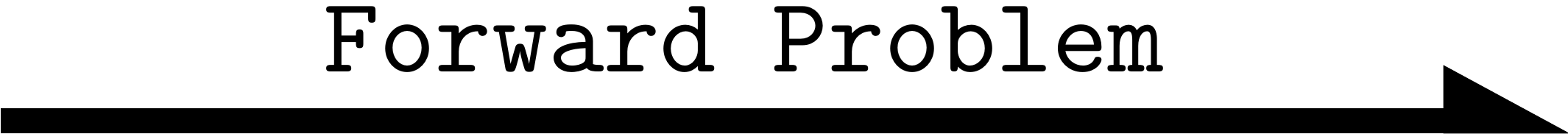}
    \end{subfigure}
    \begin{subfigure}[b]{0.31\linewidth}
        \includegraphics[width=\textwidth]{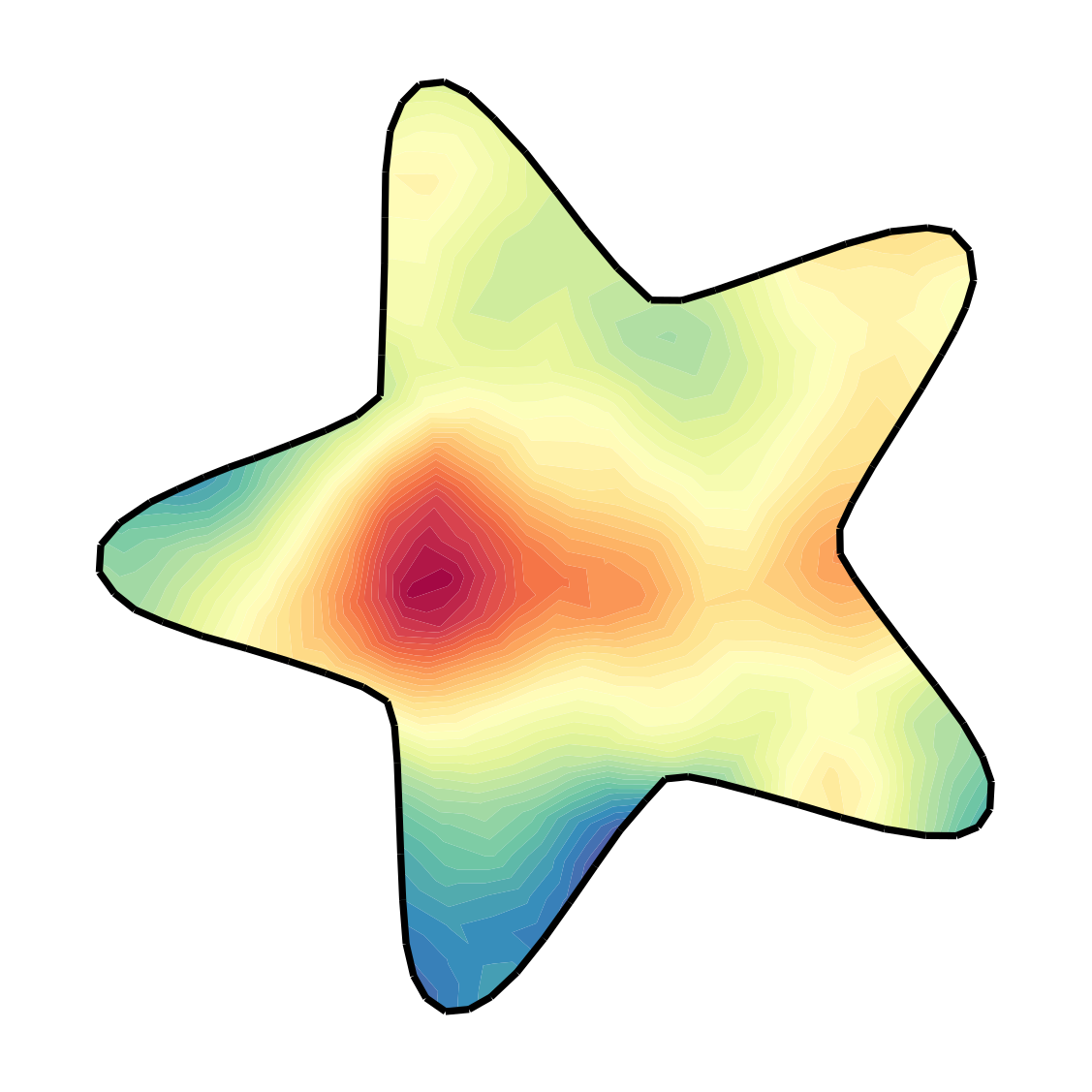}
        \caption{Input source}
        \label{fig:pde_as_smoothers_a}
    \end{subfigure}
    \begin{subfigure}[b]{0.32\linewidth}
        \includegraphics[width=\textwidth]{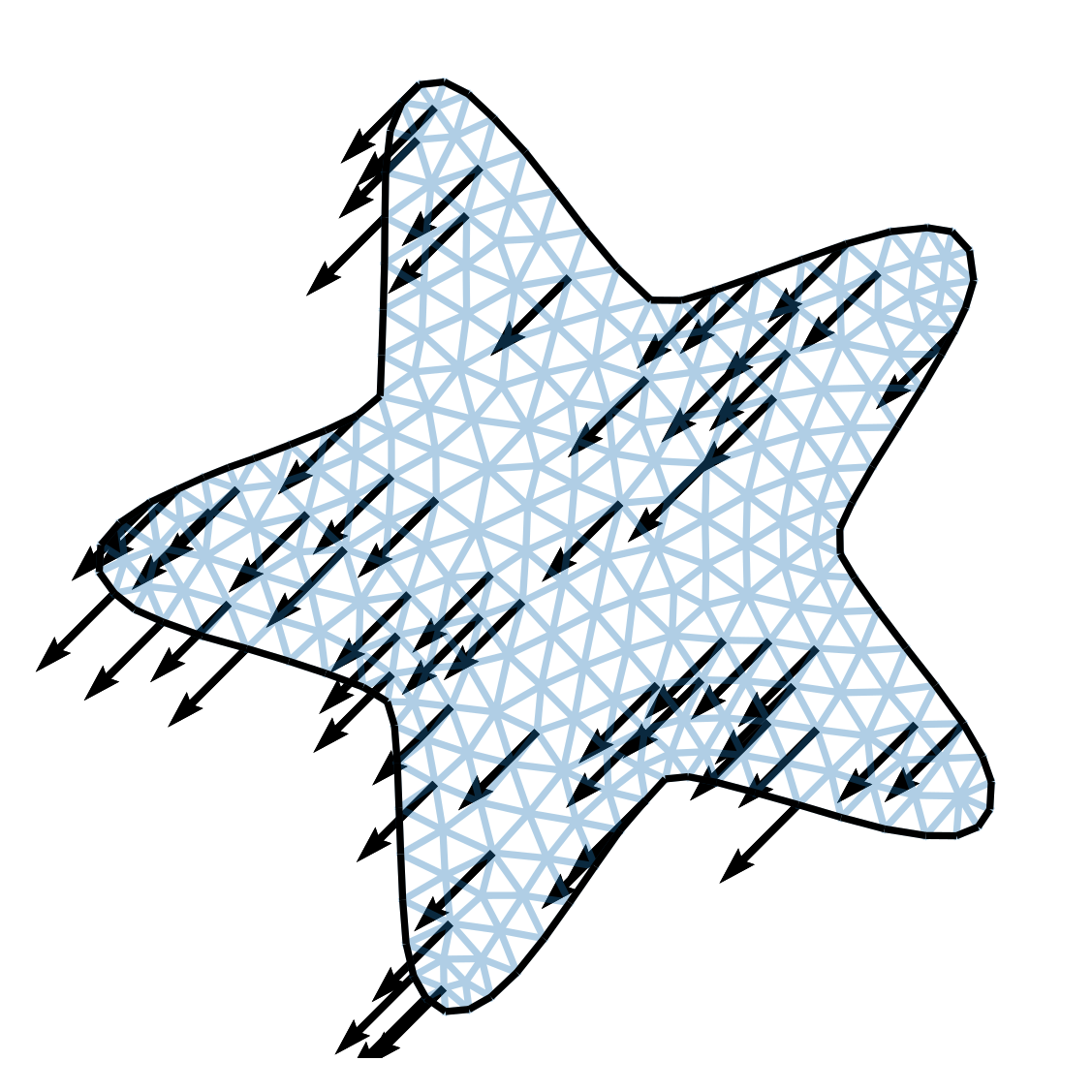}
        \caption{\centering Transport vector field}
        \label{fig:pde_as_smoothers_b}
    \end{subfigure}
    \begin{subfigure}[b]{0.31\linewidth}
        \includegraphics[width=\textwidth]{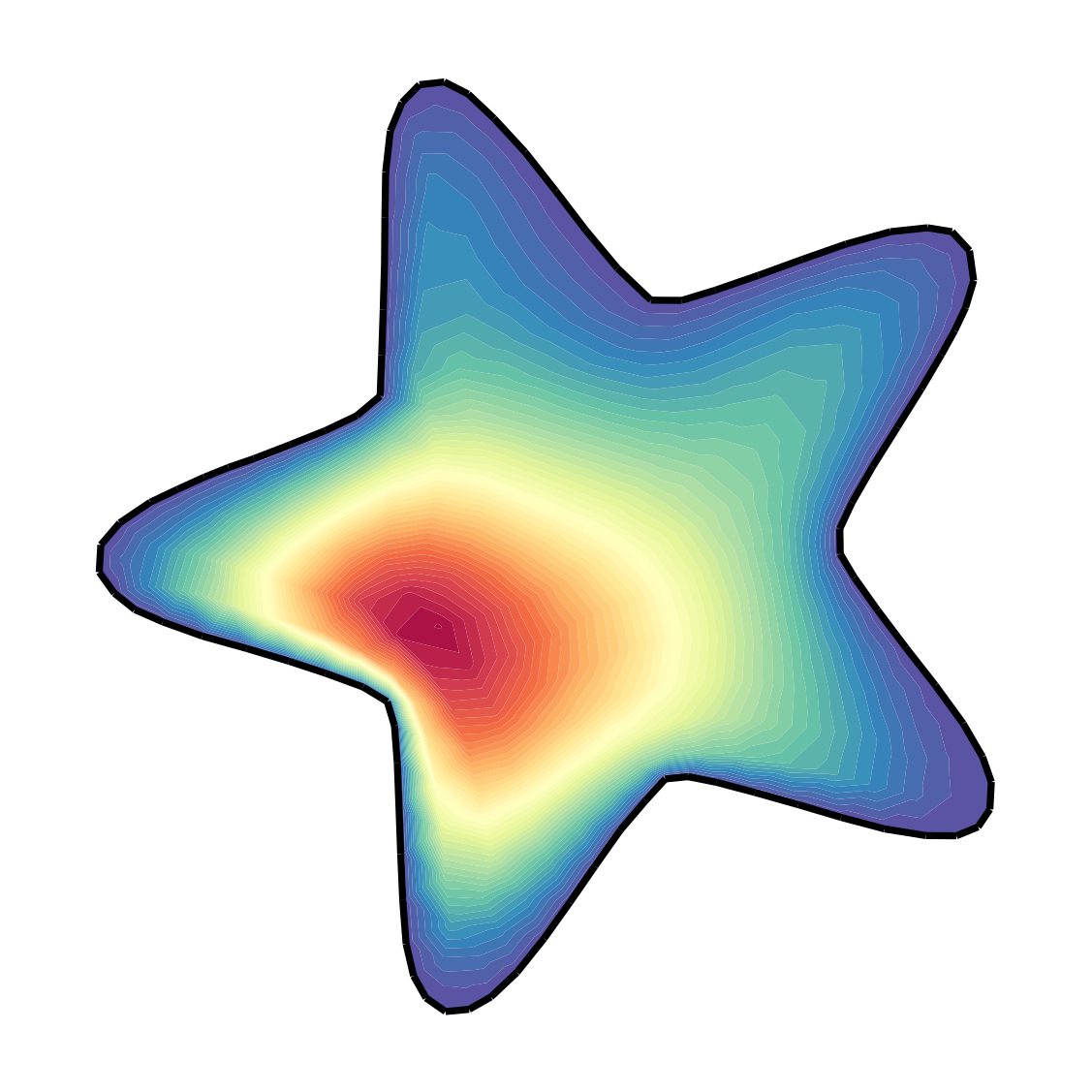}
        \caption{Output field}
        \label{fig:pde_as_smoothers_c}
    \end{subfigure}
    \vskip 0.1in
    \stepcounter{subfigure}%
    \begin{subfigure}[b]{0.31\linewidth}
        \includegraphics[width=0.9\textwidth]{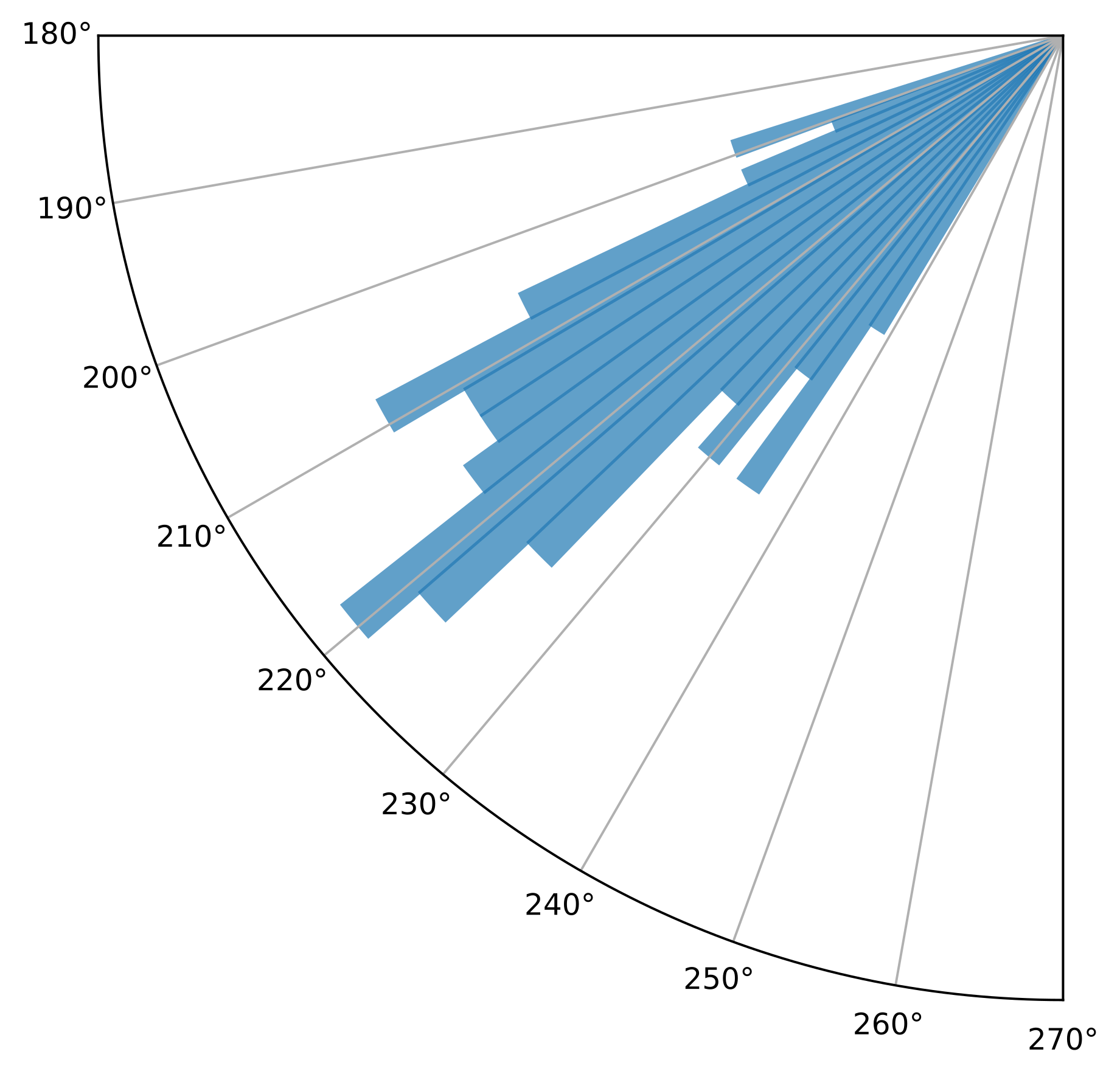}
        \caption{Posterior}
        \label{fig:pde_as_smoothers_d}
    \end{subfigure}
    \hskip 0.1in
    \addtocounter{subfigure}{-2}%
    \begin{subfigure}[b]{0.31\linewidth}
        \includegraphics[width=\textwidth]{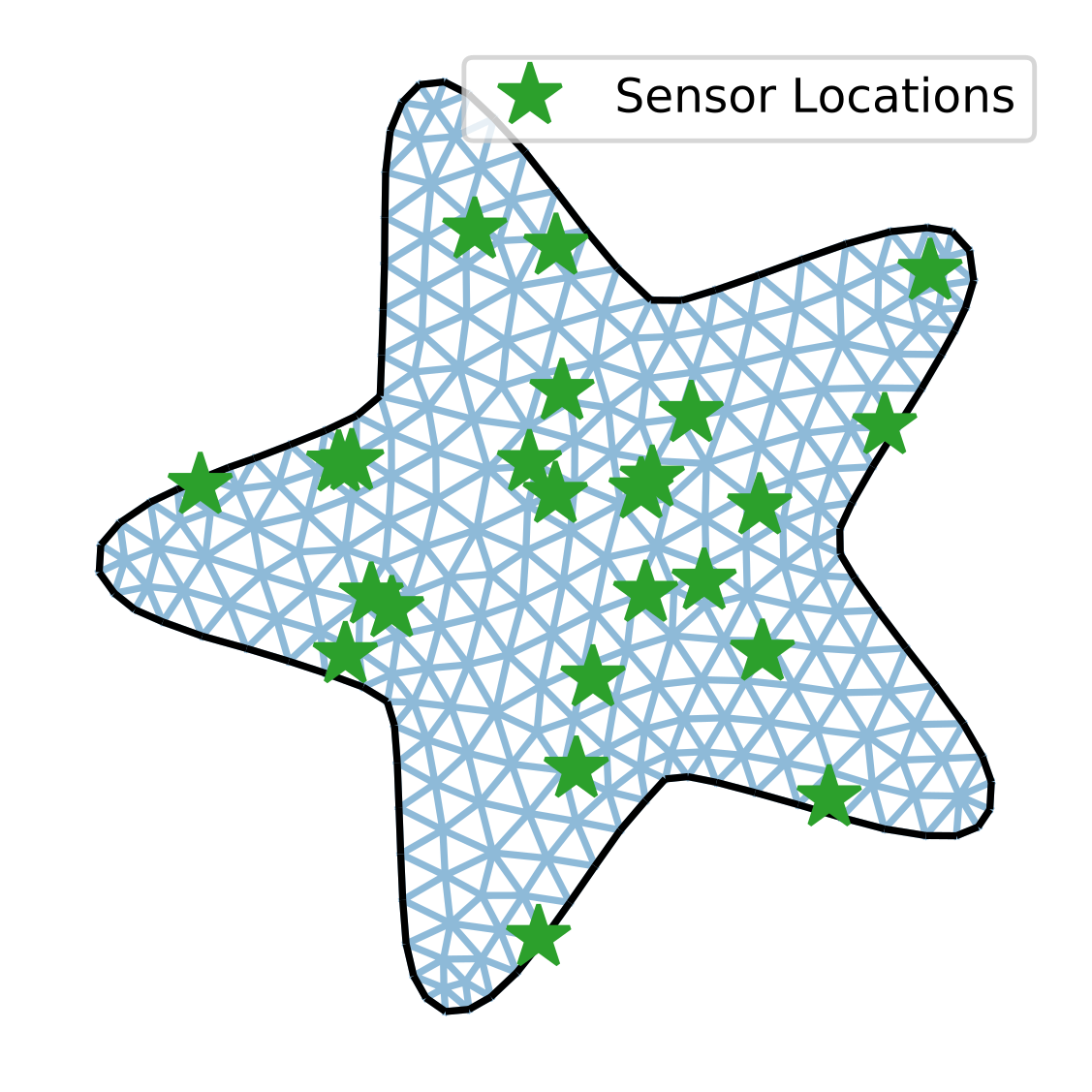}
        \caption{Sensors}
    \end{subfigure}
    \begin{subfigure}[b]{\linewidth}
    \vskip 0.1in
    \centering
        \includegraphics[width=.5\textwidth]{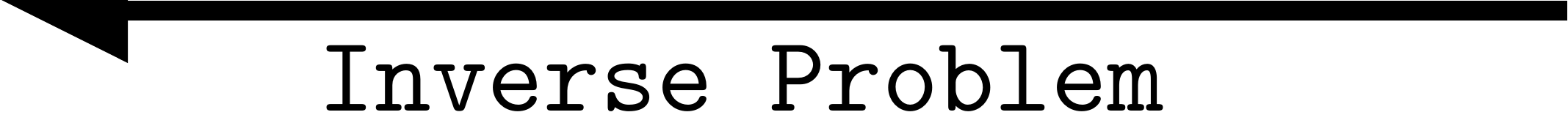}
    \end{subfigure}
    \caption{In the forward problem a noisy source (a), for example a pollutant, 
    is convolved with some transporting vector field, (b), such as wind direction,
    to produce a smoothed output in a bounded domain (c). The Bayesian
    inverse problem uses information from a finite number of sensor locations (d)
    to recover the posterior distribution of the parameters of the 
    forward map, in this instance the posterior direction vectors (e).}
\label{fig:pde_as_smoothers}
\end{wrapfigure}

Many important problems in science and engineering take the 
form of an \emph{inverse problem} \citep{stuart_2010}. The typical
task is to conduct inference over a parameter set indexing a 
mechanistic model, often presented as a partial differential equation (PDE), 
given observations. PDE based models allow one to specify 
and control complex interactions between inputs, ouputs and domain 
properties. Increasingly attention has switched to the potential of the
forward map to act as a mechanistically inspired convolution, Fig. \ref{fig:pde_as_smoothers}, 
with some of these features successfully exploited 
by \citep{alvarez_2009, arno, de2018deep, wang2019}.

However, to leverage the full power of PDEs for inference, 
one must usually solve the forward problem, typically a nonlinear and 
expensive operation, making both exact and approximate posterior inference 
challenging. This has led to development of simulation based methods
\citep{cranmer}, gradient matching methods avoiding 
numerical integration \citep{calderhead_2009, gorbach}, and recently 
fast approximations to the likelihood \citep{kersting1} based on
probabilistic numerics \citep{oates, hennig}. Common to these 
approaches is an assumption that a parameterised mechanistic model
is the \emph{correct} model, however for many real world applications
such certainty is difficult to countenance, necessitating methods which
balance the structure of mechanistic models, with the flexibility
to handle incomplete information.

Given these challenges it is natural to attempt to replace the 
forward map by some suitable family of approximating functions.
Gaussian process (GP) surrogates offer a consistent probabilistic
framework complete with uncertainty quantification 
\citep{alvarez_2009, raissi_17, girolami_2019}, but are difficult
to extend to nonlinear problems. Conversely, while deep 
generative methods easily handle nonlinear latent variables, it
is harder to guarantee that the forward map is constrained by 
the underlying physics. Supervised deep learning 
methods have imposed this constraint by applying the PDE
operator point-wise as an additional regularisation term 
\citep{Raissi, SIRIGNANO20181339, Berg, RAISSI2019686, Yazdani865063} 
which, while allowing for efficient inference, can suffer 
when data is noisy \citep{voit, chou, dattner2015}. Furthermore,
extending these regularisation approaches to generative models is
challenging, and while progress has been made on phyiscally 
constrained generative models \cite{zhu}, and structure preserving architectures \citep{pmlr-v97-cohen19d, NIPS2019_9580}, demonstrating the flexible integration of more general networks 
with only partially known physics on real world problems
remains challenging.

We address these open problems by utilising the weak-, 
or variational, form of the PDE problem as an 
inter-domain procedure \citep{interdomain} 
which augments learning with a fine-scale discretisation of the
mechanistic model to perform regularisation on the dual problem. 
In Section \ref{sec:struct_preserving_VI} we show that a consistent description of 
our generative model begins from a dual-space relaxation
of the mechanistic model which we embed inside a constrained optimisation problem. Unlike
point-wise methods, we require global regularisation,
necessitating the development in Sec. 
\ref{sec:fwd_model} of a 
local approximation, enabling 
substantial computational gains whilst maintaining 
accuracy on the inverse problem. We then empirically demonstrate 
the accuracy and efficiency of our method in Sec. \ref{sec:experiments} before concluding with a discussion and suggested future research

In summary, we demonstrate the embedding of partial 
physical knowledge into general decoders by relaxing 
a mechanistic model before using a constrained optimisation
framework to produce a coherent generative structure supervised 
by a PDE inverse problem. To further enable regularisation 
by fine-scaled discretisations of the guiding PDE we also 
introduce a 
novel approximation over local meshes 
for efficient 
training. With these steps complete we show the 
resulting models' ability to accelerate the classical BIP \emph{or}
to regularise general DL methods when prior
physical knowledge is incomplete. We first assess the 
accuracy of our approach to the BIP using synthetic data; then 
demonstrate the applicability of our methods 
as physically informed plug-in enhancements for modelling 
real-world problems in heavy-metal contamination and water-resource management.

\section{Bayesian inverse problems}
\label{sec:review}
The objective when solving a Bayesian inverse problem (BIP) \citep{stuart_2010} is to recover the 
posterior distribution of latent parameters, $\bz$,
which parameterise a \emph{forward map}, $G[\bz]$, from 
a spatial domain $\Omega \subset \mathbb{R}^D$ to
observations.
We denote the output from the forward map at 
location $\bx$ by the \emph{field variable} 
$u(\bx) = G[\bz](\bx)$, here $G$ will arise
as the solution to some mechanistic problem.
The data for the inverse problem consists of a finite set of noisy 
observations $\mathbf{y} = \{y(\bx)_n\}_{n=1}^N$
observed at spatial locations $\mathbf{X} = \{ \bx_n \}_{n=1}^{N}$, 
where $\mathbf{x}_n \in \Omega \subset \mathbb{R}^D$, with
the field variable acting as a deterministic link from 
the latent variables to observations in these locations through a 
likelihood $p(\by \mid \bu)$, and the target posterior is
$p(\bz \mid \by) \propto p(\by \mid \bu = G[\bz](\bx))p(\bz).$
Inference up to an unknown normalising constant is 
standard, however the forward map is typically unavailable 
in closed form requiring expensive numerical methods. In this 
work we embed the BIP inside a constrained optimisation 
framework, following the optimisation view of Bayesian inference
\citep{zellner_1988} that recovers the posterior as:

\begin{empheq}{alignat=2}
    \qquad \argmin_{q(\theta) \in \mathcal{P}(\Theta)} \quad \mathbb{E}_{q(\theta)}
    \left[-\log p(\by \mid \theta) \right]
    + \KL(q(\theta)\,\|\, p(\theta) ),
    \label{eq:posterior_optim_problem}
\end{empheq}

where $\KL(q\,\|\,p)$ denotes the KL-divergence 
between distributions with densities $q$ and $p$, 
and $\mathcal{P}(\Theta)$ denotes the set of all distributions 
on the general parameter spaces $\theta \in \Theta$.
Since the forward operator $G[\bz]$ is deterministic 
we rewrite this formally as the equivalent constrained problem
\begin{empheq}{alignat=2}
    \argmin_{q(\bu, \bz) \in \mathcal{P}(V \times \mathcal{Z})}
    \quad &\mathbb{E}_{q(\bu, \bz)}\left[-\log p(\by \mid \bu) \right]
     + \KL(q(\bu, \bz)\,\|\,p(\bu, \bz) ) 
    \quad \mbox{s. t. } p(\bu \mid \bz) = \delta_{G[\bz]}(\bu),
\label{eq:const_posterior_optim_problem}
\end{empheq}
where $\delta_{G(\bz)}(\bu)$ denotes the degenerate Dirac distribution
centred on the deterministic solution and $V$ is a function
space in which we assume the solution of the forward problem
to exist, we refer to this as the space of
\emph{trial functions} \citep{brezis}.
In this work we initially relax the constrained problem 
\eqref{eq:const_posterior_optim_problem}, 
before showing that the structure can be recovered, allowing us 
to introduce variational methods 
respecting the generative structure of the model, but avoiding
expensive evaluations of the forward map.

\subsection{Partial differential equations and finite element discretisation}
\label{sec:fem}
We focus on a particular instance of the BIP by examining forward maps 
which arise as the implicit solutions of PDEs. These models are a 
bedrock of modern science and engineering and they are presented 
in the form
\begin{align}
    \mathcal{L}[\bz] u(\bx) = f(\bx) \mbox{ on } \Omega, \qquad 
    u(\bx) = g(\bx) \mbox{ for } \mathbf{x} \in \partial\Omega,
\label{eq:gen_pde}
\end{align}
where $\mathcal{L}[\bz]$ is a differential operator 
parameterised by
$\bz$.\footnote{The forward map $G$ now 
arises as the implicit solution to the PDE
\eqref{eq:gen_pde}}
The source, $f(\mathbf{x})$, represents domain wide 
inputs, while the boundary condition $\mathbf{g}(\bx)$ describes 
inputs acting only through the boundary, $\partial \Omega$. 
These models are complicated because \eqref{eq:gen_pde} 
only defines the forward map implicitly, but can represent 
a diverse range of interesting dynamics. We can 
illustrate some of this richness 
by considering
\begin{align}
    \mathcal{L}[\bz] (\cdot) = -\nabla (a(\bx, \bz) \nabla (\cdot) )
    + \boldsymbol{\tau}(\bx, \bz) \cdot \nabla (\cdot)  \label{eq:transport}
\end{align}
where $a(\bx)$ is the \emph{diffusion coefficient} and
$\boldsymbol{\tau}(\bx)$ is referred to as the
\emph{transport vector field}.
We visualise the action of this 
operator in Fig. \ref{fig:pde_as_smoothers}; in this instance we have
an input source, for example pollutants represented in 
Fig. \ref{fig:pde_as_smoothers_a}. The operator is parameterised by a 
transporting vector
field $\boldsymbol{\tau}(\bx)$, such as the prevailing 
wind-direction, visualised in Fig. \ref{fig:pde_as_smoothers_b}.
The forward map can be viewed as a 
structured convolution, where pollutants are smoothed 
by the diffusion parameter $a(\bx)$, and transported 
by $\boldsymbol{\tau}(\bx)$ until they encounter the boundary. 
The solution $u(\bx)$ is represented in Fig. 
\ref{fig:pde_as_smoothers_c} in which we can clearly observe 
diffusion, transport and boundary effects.

Since the forward map is unavailable in closed 
form it becomes necessary to solve 
the PDE numerically, and the finite element method \citep{brenner} (FEM) is 
a powerful method for doing so. To illustrate our review we shall discuss the 
Laplace operator, $\Delta = \nabla^2$, which leads to a special case of \eqref{eq:transport} as
\begin{align}
    -\Delta u(\bx) = f(\bx)  \quad \mbox{ for } \bx \in \Omega, \qquad 
    u(\bx) = g(\bx) \quad \mbox{on } \partial \Omega. \label{eq:poisson}
\end{align} 
To discretise one introduces a second function space, denoted $\hat{V}$,
of \emph{test functions} $v \in \hat{V}$, 
then multiplies the classical form \eqref{eq:poisson}
by $v$ and integrates to give the \emph{weak form} of the classical
problem
\begin{align}
    \int_{\Omega} \nabla u(\bx) \cdot \nabla v(\bx) d\bx = 
    \int_{\Omega} f(\bx) v(\bx) d\bx - 
    \int_{\partial \Omega} g(\mathbf{s}) v(\mathbf{s}) ds. \label{eq:finite_var_problem}
\end{align}
The variational problem is now: find solution $u \in V$ 
such that \eqref{eq:finite_var_problem} holds for any 
test function $v \in \hat{V}$. 
To numerically implement this idea using the FEM method
one decomposes $\Omega$ into a collection of disjoint elements, 
and specifies a finite dimensional set of basis functions 
$V = \mathrm{span}\{\phi_i \}$ and 
$\hat{V} = \mathrm{span}\{\hat{\phi}_i\}$ for the test and
trial space
respectively, which are completely determined by their
values on the nodes $\{\bar{\mathbf{x}}_j \}_{j=1}^{N_{\text{nodes}}}$
of the mesh, see \citep{reddy} for a full review of how this
process is implemented. Having specified the basis, one searches for 
solutions with a representation $u_n(\bx) = \sum_{j=1}^N (\boldsymbol{\xi})^{u}_j \phi_j(\bx)$, where 
$\boldsymbol{\xi}^u \in \mathbb{R}^{\Nnodes}$ are unknown coefficients. In matrix-vector notation we have
$\mathbf{A} \boldsymbol{\xi}^u = \bff$, where the 
\emph{stiffness matrix} and \emph{load vector} are given by
\begin{align}
    (\mathbf{A})_{ij} = \int_{\Omega} \nabla \phi_j(\bx) \cdot \nabla \hat{\phi}_i(\bx) dx 
    \qquad 
    (\bff)_i = \int_{\Omega} f(\bx) \hat{\phi}_i(\bx) dx 
    - \int_{\partial\Omega} g(\mathbf{s}) \hat{\phi_i}(\mathbf{s}) d\mathbf{s}.
    \label{eq:stiffness_and_load_vector}
\end{align}
The use of nodal basis functions guarantees that 
$(\mathbf{A})_{ij}$ is zero whenever $\bar{\mathbf{x}}_{i}$ 
and $\bar{\mathbf{x}}_j$ are not in adjacent elements.
Following the same procedure we can discretise 
\eqref{eq:transport} as
\begin{align}
&(\mathbf{L}[\mathbf{z}])_{ij} = \int_{\Omega} a(\bx, \bz) \nabla \phi_j(\bx)
\cdot \nabla \hat{\phi}_i(\bx) dx 
+ \int_{\Omega} \boldsymbol{\tau}(\bx, \bz) \cdot \nabla \phi_j(\bx) 
\hat{\phi}_i (\bx)dx. \label{eq:transport_assembly}
\end{align}
We stress that while in principle these quadratures are 
defined over the entire domain, in practice owing to the 
sparsity of the nodal basis functions it is only necessary 
to perform this quadrature locally over adjacent elements. The solution of a PDE by 
the FEM therefore requires two steps
\begin{itemize}
    \item[(i)] [\texttt{Assembly}] The process of pushing forward the 
    latent process $\mathbf{z} \mapsto \mathbf{L}[\mathbf{z}]$ through
    the quadrature \eqref{eq:transport_assembly}. This is a 
    $\mathcal{O}(N_{\text{elements}})$ embedding into a sparse matrix.
    \item[(ii)] [\texttt{Solve}] Inverting to solve $\boldsymbol{\xi}^u
    = \mathbf{L}[\mathbf{z}]^{-1}\bff$. Direct application of this 
    is $\mathcal{O}(N_{\text{nodes}}^3)$.
\end{itemize}
To solve the BIP using the FEM it would be necessary to perform 
repeated calls to \texttt{Assembly} and \texttt{Solve}. In this work we shall only ever need to perform the cheap 
\texttt{Assembly} operation.

\section{Structure-preserving Constrained VI}
\label{sec:struct_preserving_VI}
If we are going to circumvent the true forward map, but still 
encode the physical structure, then
we must introduce an alternative means of ensuring the mechanism
is embedded into the inferential process.
In this section we show
how we can do that through a VI framework which is augmented with the discretised weak-form that in turn encodes the
mechanistic model. We first relax the inherent degeneracy of the objective problem 
\eqref{eq:const_posterior_optim_problem}, by introducing a one 
parameter family of approximating conditional densities, $p_{\epsilon}(\bu \mid \bz)$. Our goal is to now specify
such a $p_{\epsilon}$ so as to replace the original
objective function by
\begin{align}
    F_{\epsilon} = \mathbb{E}_{q(\bu, \bz)}\left[ 
    - \log p(\by \mid \mathbf{u}) \right]
    + \KL(q(\bz) \| p(\bz)) 
    + \mathbb{E}_{q(\bz)}\left[ \KL(q(\bu \mid \bz) 
    \| p_{\epsilon}(\bu \mid \bz)) \right]. \label{eq:relaxed_problem}
\end{align}

\subsection{Relaxing the VI problem}
\label{sec:relaxing_the_vi_prob}
One immediate way of achieving a relaxation 
that continues to respect the underlying PDE structure
is to replace the original model \eqref{eq:gen_pde}
by
\begin{align}
    \mathcal{L}[\bz] u(\bx) = f(\bx) + \epsilon w(\bx),
    \label{eq:perturb}
\end{align}
where $w(\bx)$ is an independent Gaussian process perturbation, and
$\epsilon > 0$ is a scaling parameter. Using the FEM described in
Section \ref{sec:review} to discretise we may write
\begin{align}
    p_{\epsilon}(\bu \mid \bz) = 
    \mathcal{N}( \bu \mid \mathbf{L}[\bz]^{-1}\mathbf{f}, 
    \epsilon^2 (\mathbf{L}[\bz]^{\top} \mathbf{D} 
    \mathbf{L}[\bz])^{-1}) 
    \label{eq:measure_in_greens_form}
\end{align}
where $\mathbf{D}$ is the precision matrix of the
process $w(\bx)$ after projection onto the test space.
However, it remains unclear how one 
should specify the covariance operator of $w(\bx)$. In fact we shall 
see below that in the limit this choice would not matter, but for
practical implementations it does.

Our approach is motivated by \cite{ito_kunisch_1990} who
note that if $u$ is an element of the Sobolev space $H_{0}^1$, 
then through the weak-form \eqref{eq:finite_var_problem}
a pair $(u, \bz)$ determines an element, which we denote by 
$\varphi_{u, \bz}$, of the dual space $H^{-1}$. Ideally this
should be small, so we place a Gaussian measure on
$H^{-1}$ and then use the Riesz-Fr\'echet isomorphism and
properties of the $H^{1}_0$ norm to construct a Gaussian measure
\begin{align*}
    \exp\left\{
    \frac{-1}{2\epsilon^2} \| \varphi_{u, \bz} \|^2_{H^{-1}}
    \right\}
    = 
    \exp\left\{
\frac{-1}{2\epsilon^2} \| (-\Delta^{-1}) \varphi_{u, \bz}
    \|^2_{H_{0}^1}    
    \right\}
    = 
    \exp\left\{
    \frac{-1}{2\epsilon^2} \langle 
    \varphi_{u, \bz}, (-\Delta)^{-1} \varphi_{u, \bz}
    \rangle_{L^2(\Omega)}   
    \right\}
\end{align*}
now replacing the dual space element $\varphi_{u, z}$ by the discretised 
weak-form $\mathbf{L}[\bz]\mathbf{u} - \mathbf{f}$, and $(-\Delta)^{-1}$ 
by the inverse stiffness matrix, then we arrive at the density
\begin{align*}
    p_{\epsilon}(\bu \mid \bz) \propto \exp\left\{ 
    -\frac{1}{2\epsilon^2}
    \left( \mathbf{L}[\bz]\mathbf{u} - \mathbf{f} \right)^{\top}
    \mathbf{A}^{-1}
    \left( \mathbf{L}[\bz]\mathbf{u} - \mathbf{f} \right)
    \right\}
\end{align*}
which after rearranging is precisely \eqref{eq:measure_in_greens_form}, 
with the precision matrix given by the stiffness matrix 
\eqref{eq:stiffness_and_load_vector} from the FEM applied to the 
Poisson problem \eqref{eq:poisson}, that is $\mathbf{D}
= \mathbf{A}^{-1}$.
Intuitively, our relaxed model 
allows pairs 
$(u, \bz)$ which do not exactly satisfy the mechanistic model, and 
then penalises these deviations from exact solution pairs according 
to a Gaussian measure on the dual space, which we interpret as a 
space of ``approximate mechanisms''
While the idea of perturbing \eqref{eq:perturb}
to create a relaxed problem is not a new one, and is fundamental
to emulator based methods \citep{kennedy_ohagan_2001}, our 
dual-space relaxation leads to a somewhat 
counter-intuitive perturbing process. For instance, a
natural choice of perturbation to \eqref{eq:perturb}
would be the stochastic PDEs considered in 
\citep{Lindgren}, characterised by dense covariances
but sparse precisions 
allowing for efficient inference. Instead we are led to consider the reverse of this, 
and this will have important implications
which we revisit in Sec. \ref{sec:fwd_model}.

\subsection{Re-constraining the VI problem}
\begin{figure}[t!]
\begin{subfigure}[t]{.5\textwidth}
  \centering
    \includegraphics[height=3.8cm]{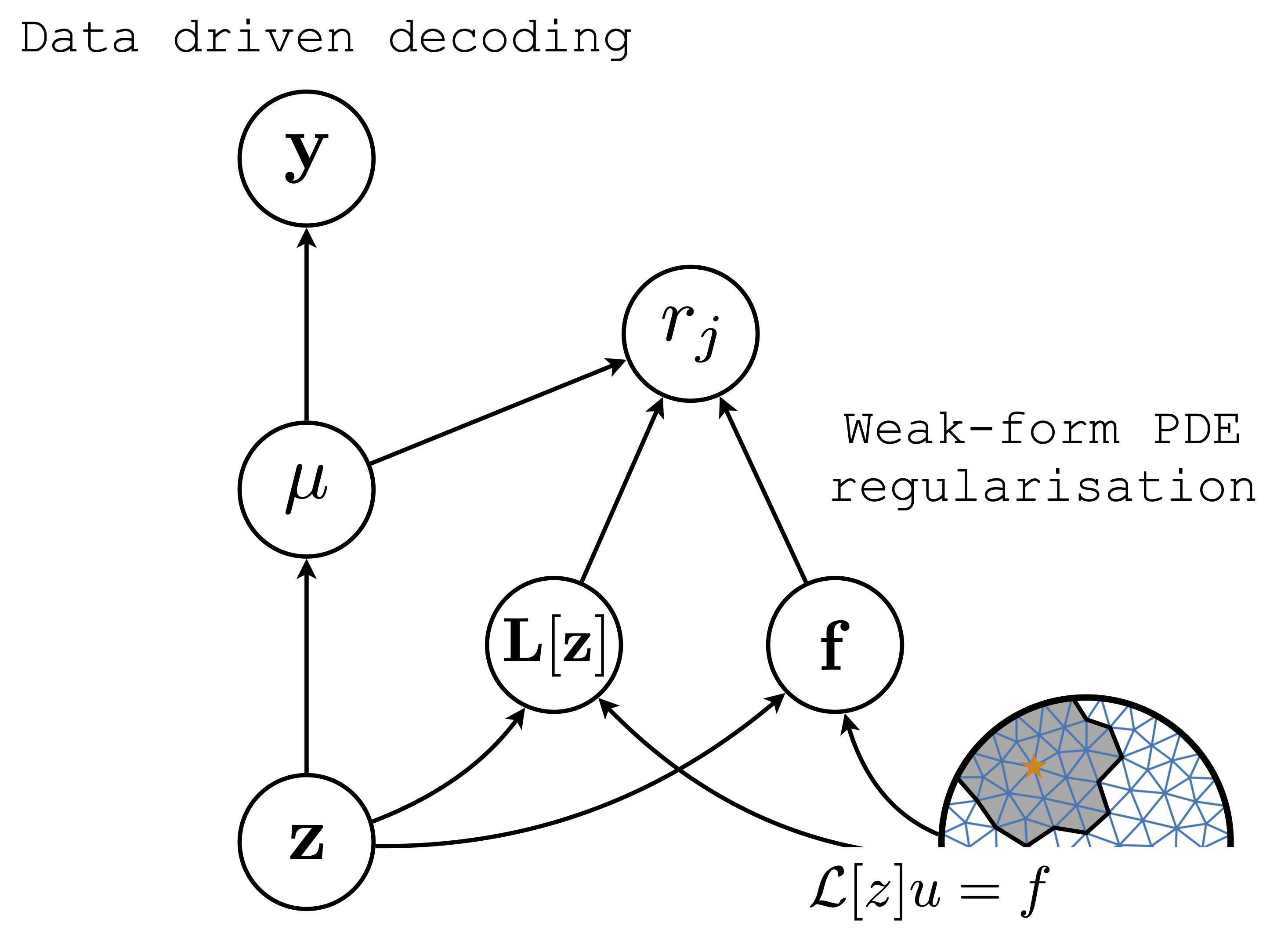}
  \caption{PDE regularised decoder} \label{fig:pde_reg_decoder}
 \end{subfigure}\hfill
 \begin{subfigure}[t]{.45\textwidth}
    \includegraphics[height=3.8cm]{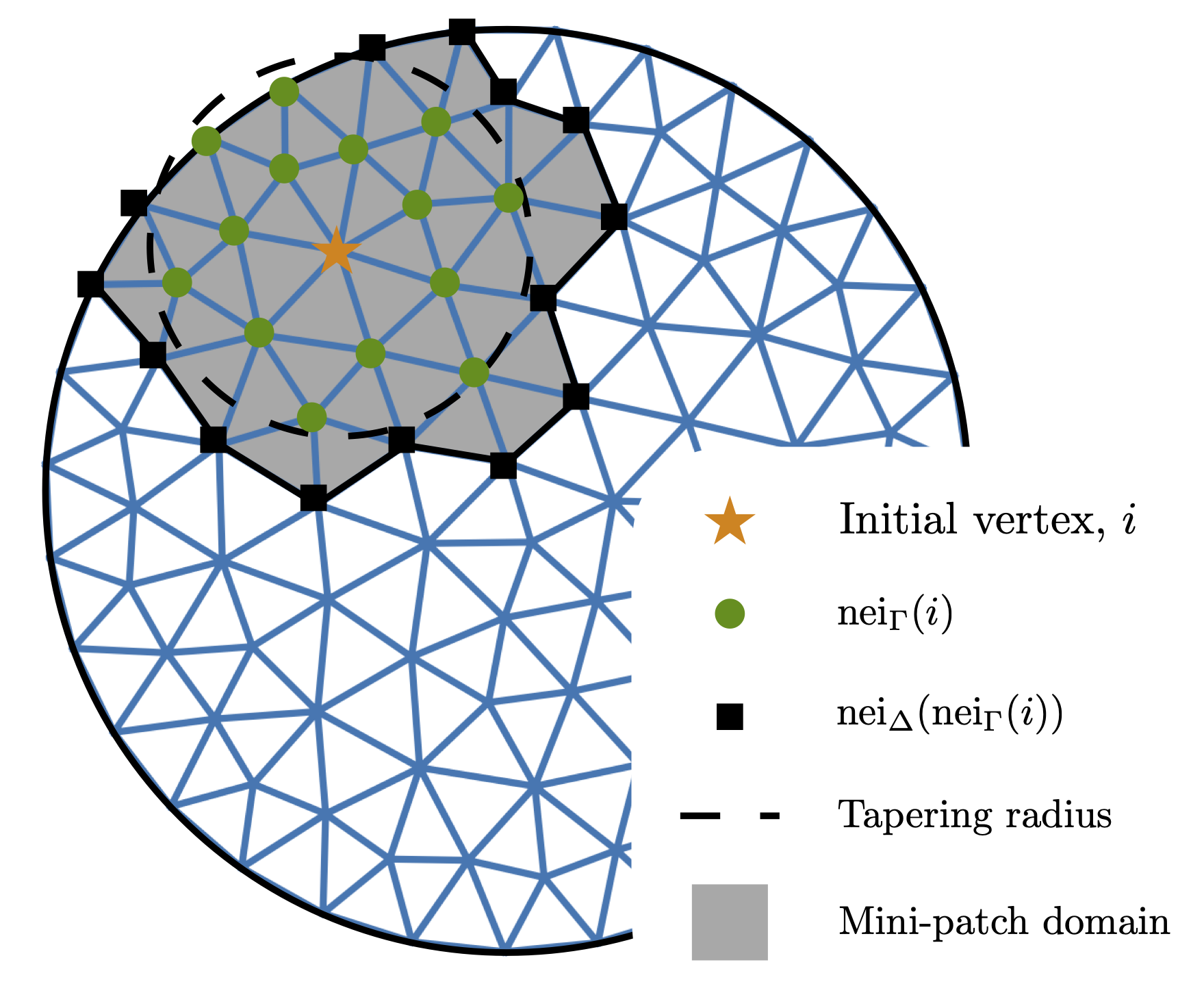}
    \caption{Local mini-patch approximation}
    \label{fig:minipatch}
\end{subfigure}
\caption{(a) Regularisation of a standard decoder network
with the weak form PDE structure. (b) Illustration of our tapering approach to construct local regularisers 
for computational efficiency.}
\end{figure}

We now demonstrate how our approximation allow us to tackle 
the variational problem for the relaxation \eqref{eq:relaxed_problem},
and to examine the consequences of taking the limit 
$\epsilon \rightarrow 0$. Our variational family
is distributions with conditionals of the form
\begin{align}
    q(\bu \mid \bz) = \mathcal{N}(\bu \mid \mu(\bx, \bz), 
    \epsilon^2 (\mathbf{L}[\bz]^{\top}\mathbf{A}^{-1}\mathbf{L}[\bz])^{-1}),
    \label{eq:approximating_var_cond}
\end{align}
where $\mu(\bx, \bz)$ is some free function of the spatial coordinates
and latent parameters. We shall assume 
that the family of approximating functions has 
sufficient capacity to express the forward map, and therefore 
the exact conditional is contained within our family of approximating
distributions. After integrating over $q(\bu \mid \bz)$ then
\eqref{eq:relaxed_problem} becomes
\begin{align}
    F_{\epsilon} = \mathbb{E}_{\bu \sim q(\bu, \bz)}\left[ - \log p(\by \mid \bu) \right] + \KL(q(\bz) \| p(\bz)) 
    + \frac{1}{2\epsilon^2}
    \mathbb{E}_{q(\bz)}\left[
    \left\| \mathbf{L}[\bz]\mu(\bx, \bz) - \mathbf{f}\right\|_{\mathbf{A}}^2
    \right], \label{eq:cvi_sub_prob}
\end{align}
where $\|\mathbf{w}\|_{\mathbf{A}}$ denotes the quadratic form 
$\mathbf{w}^{\top}\mathbf{A}^{-1}\mathbf{w}$, note this has
allowed us to avoid evaluating the covariance matrix in \eqref{eq:approximating_var_cond},
which requires the prohibitive \texttt{Solve}. The first two terms 
of \eqref{eq:cvi_sub_prob} are standard, and indeed are 
reminiscent of the decoding network of a variational 
autoencoder \citep{kingma, rezende14}
and we expand on that connection in the supplement. 
Less standard is the final term which provides strong
mechanistic regularisation of the encoder network. To see this, let 
$\{ \bz_i \}_{i=1}^{M}$ be a collection of independent samples from
$q(\bz)$, then we can obtain the MC approximation of 
\eqref{eq:cvi_sub_prob} using
\begin{align}
    \mathbb{E}_{\bz \sim q(\bz)}\left[ 
    \|\mathbf{L}[\bz]\mu(\bx, \bz) - \mathbf{f} \|_{\bA}^2
    \right]
    \approx \frac{1}{M} \sum_{i=1}^M
    \| \mathbf{L}[\bz_i]\mu(\bx, \bz_i) - \mathbf{f} \|_{\bA}^2.
    \label{eq:mc_reg_term}
\end{align}
Defining the scalar variable $r(\bz) \stackrel{\Delta}{=} \| \mathbf{L}[\bz]\mu(\bz) - \mathbf{f} \|_{\bA}$, then we can approximate \eqref{eq:cvi_sub_prob} as 
\begin{align}
    F_{\epsilon} \approx \mathbb{E}_{\bu \sim q(\bu, \bz)}
    \left[- \log p(\by \mid \bu) \right] 
    + \KL(q(\bz) \| p(\bz))
    + \frac{1}{2\epsilon^2 M} 
    \sum_{i=1}^M r(\bz_i)^2.
    \label{eq:MC_cvi_sub_prob}
\end{align}
One can take the $\epsilon \rightarrow 0$ limit and recognise \eqref{eq:MC_cvi_sub_prob} as a quadratic penalty representation 
\citep{nocedal_wright} of
\begin{subequations}
\begin{empheq}{alignat=2}
    &\argmin_{q(\bz) \in \mathcal{Q},\, \mu \in V}
    \mathbb{E}_{\bz \sim q(\bz)}
    \left[- \log p(\by \mid \mu(\bz, \bx)) \right] 
    + \KL(q(\bz) \| p(\bz)) 
    \label{eq:eq_const_posterior_optim_problem_objfunc}
    \\
    &\mbox{subject to } r(\mathbf{z}_i)=0, \text{ for all finite samples
    $\{\mathbf{z}_i\}_{i=1}^{M}$ from $q(\bz)$} \label{eq:fem_constrained}
\end{empheq} \label{eq:new_cvi_problem}
\end{subequations} 
Since the sample was arbitrary we can also interpret \eqref{eq:fem_constrained} 
as requiring that the random variable $r = r(\bz)$ is $\delta$-distributed. 
This strong constraint is the VI counterpart to the motivating
problem \eqref{eq:const_posterior_optim_problem}. Returning to the analogy with 
the VAE, by removing the need to exactly solve the PDE we have allowed a high 
capacity free-form map to best express the data, however we reintroduce the 
mechanistic structure as a strong encoding term, not from the data to the latent 
space, but \textit{from the latent space to a point-mass determined 
by the mechanistic structure}. This allows us to recast the 
augmentation of the model with a mechanistic
structure, as a (pseudo)-data augmentation scheme with the new output-target pairs given by $\{(\mathbf{z}_m, 0) \}_{m=1}^M$, 
and auxillary data taking the form of a mechanistic model and its dense FEM discretisation; this resulting decoder is 
displayed in Fig. \ref{fig:pde_reg_decoder}.

After passing to the limit in \eqref{eq:new_cvi_problem} the
dependence on the additive perturbation, has disappeared. If one 
has complete confidence in the specified model this limiting problem may be desirable, 
however the imposition of hard constraints in ML architectures remains under-developed, 
\citep{marquezneila}. Of more philosophical import; it is unlikely we would ever posses 
such absolute certainty. The relaxation we introduce quantifies uncertainty in the 
forward map, accounting for \textit{unknown physics} and the over-simplifications that occur when 
deriving mathematical models of complex processes.

\section{Fast forward approximations}
\label{sec:fwd_model}
We have reduced the work done solving the BIP to  optimising 
a variational distribution, $q(\bz)$, and a 
variational parameter, $\mu$, penalised by applying the FEM 
weak-form constraint \eqref{eq:fem_constrained}.  It is instructive to 
consider approaches \citep{SIRIGNANO20181339, Raissi} that have
used the classical form \eqref{eq:gen_pde} to construct regularisers
\begin{align}
    \frac{1}{N_{\text{int}}}\sum_{i=1}^{N_{\text{int}}}
    \left( \mathcal{L}[\mathbf{z}](\mu(\bx_i) - f(\bx_i) \right)^2 
    + \frac{1}{N_{\text{bnd}}}\sum_{j=1}^{N_{\text{bnd}}}
    \left( \mu(\mathbf{s}_j) - g(\mathbf{s}_j) \right)^2,
    \label{eq:pointwise_pde_reg}
\end{align}
where $\{\mathbf{x}_i\}_{i=1}^{N_{\text{int}}}$ and 
$\{\mathbf{\mathbf{s}}_j\}_{j=1}^{N_{\text{bnd}}}$ are
sampled uniformly from the interior, and boundary. This allows efficient
batched gradient descent, however it is unclear how to extend 
this to produce a generative model. We could try
and modify our approach in Sec. \ref{sec:relaxing_the_vi_prob}
by viewing \eqref{eq:pointwise_pde_reg} as MC approximation of 
\begin{align}
    \exp\left\{-\frac{1}{2}  
    \| \mathcal{L}[\bz]u - f \|^2_{L^2(\Omega)} \right\},
    \label{eq:not_a_gaussian_measure}
\end{align}
however, the identity considered as a covariance operator
is not Hilbert-Schmidt, since
$\mathrm{Tr}(I) = \infty$. Therefore
\eqref{eq:not_a_gaussian_measure} is not a 
Gaussian measure preventing our Gaussian perturbation
approach from being applied. 
By originating in the 
dual-space our work addresses this 
problem; however in doing so we augmented 
with the \emph{entire} FEM structure, a 
significant increase in the computational burden 
compared to \eqref{eq:pointwise_pde_reg}. To see
this we consider an analogous MC approximation for our regulariser
\begin{align}
    \| \WeakForm \|_{\bA}^2 
    &= \sum_{i=1}^{\Nnodes} (\WeakForm)_i 
    \sum_{j \; : \; (\bA^{-1})_{ij} \neq 0} (\bA^{-1})_{ij} (\WeakForm)_j \notag \\ 
    &\stackrel{\Delta}{=} \sum_{i=1}^{\Nnodes} r_{\bA}^{(i)}(\bz) 
    \approx \frac{1}{P} \sum_{p=1}^P r_{\bA}^{(i_p)}(\bz), \qquad 
    i_1,\ldots, i_P \sim \Uniform{\Nnodes}
    \label{eq:mc_approx_to_reg}
\end{align}
where we have replaced the full outer sum, with $P \ll \Nnodes$ 
summands by uniformly sampling from the complete node set. 
Unfortunately, while $\bA$ is sparse, the precision $(\bA)^{-1}$ 
will not be, so that each $r^{(i_p)}_{\bA}$ requires a 
complete \texttt{Assembly}, preventing the efficient 
batching possible in \eqref{eq:pointwise_pde_reg}.

To handle this we propose replacing the precision with a suitably 
tapered version. This technique has been successfully applied in 
geostatistics \citep{cohn, furrer} and constructs a new matrix
$\boldsymbol{\Gamma} = (\bA)^{-1} \circ \mathbf{K}_{\text{taper}}$,
where $\mathbf{K}_{\text{taper}}$ is the Gram matrix of some tapering
function chosen so that $(\mathbf{K}_{\text{taper}})_{ij} = 0$ whenever
$\|\bar{\mathbf{x}}_i - \bar{\mathbf{x}}_j\| > \rho$ for some $\rho > 0$.
We define the $\Gamma$-neighbourhood of a vertex $i$ to be
$\text{nei}_{ \Gamma}(i) = \{ j \; : \; (\Gamma)_{ij} \neq 0 \}$, 
then approximate $r^{(i)}_{\bA}$ by $r^{(i)}_{\Gamma}$, so
reintroducing sparsity into our regularisation. Evaluation of the 
terms $r^{(i)}_{\Gamma}$ will require the values of 
spatially varying processes over the set of index points 
$\{i\} \cup \mathrm{nei}_{\Gamma}(i) \cup \mathrm{nei}_{\Delta}(\mathrm{nei}_{\Gamma}(i))$, where 
$\mathrm{nei}_\Delta$ is the natural neighbourhood structure of 
a FEM mesh. This collection of nodes implicitly defines a reduced set 
of elements over which we need to evaluate the weak form, 
$\mathbb{T}^{\text{active}},$ which we refer to as a 
\emph{mini-patch}, and display in Figure \ref{fig:minipatch}. 

Assembly over this reduced mesh will be substantially cheaper than that 
over the full mesh, and in effect corresponds to replacing our model
in Section \ref{sec:relaxing_the_vi_prob} with a misspecified covariance function \citep{stein1999, furrer}. Combined with sampling from 
the process $\bz$ we arrive at an efficient MC approximation 
to the complete penalty by independently sampling $q(\bz)$, and an initial 
vertex $i$ around which to build the mini-patch, obtaining the hierarchical 
estimate
\begin{align*}
\mathbb{E}_{\bz \sim q(\bz)}\left[ 
    \|\mathbf{L}[\bz]\mu(\bx, \bz) - \mathbf{f} \|_{\bA}^2
    \right]
    \approx \frac{1}{M}\sum_{n=1}^{M} \frac{1}{P} \sum_{p=1}^{P}
    r^{(i_p)}_{\Gamma}(\bz_n),
\quad \mathbf{z}_{n} \sim q(\bz), \quad
i_p \stackrel{\text{i.i.d}}{\sim} \mathrm{Unif}(\Nnodes)
\end{align*}
By re-sampling the patch at each step of the optimisation we
ensure this local constraint is applied everywhere so achieving global regularisation of the data-driven map.
In the next section we empirically demonstrate the accuracy of this approximation, further
study is provided in the supplement.

\section{Experiments}
\label{sec:experiments}
\begin{figure}[t]
    \centering
    \begin{subfigure}[b]{0.31\linewidth}
        \centering
        \includegraphics[width=\textwidth]{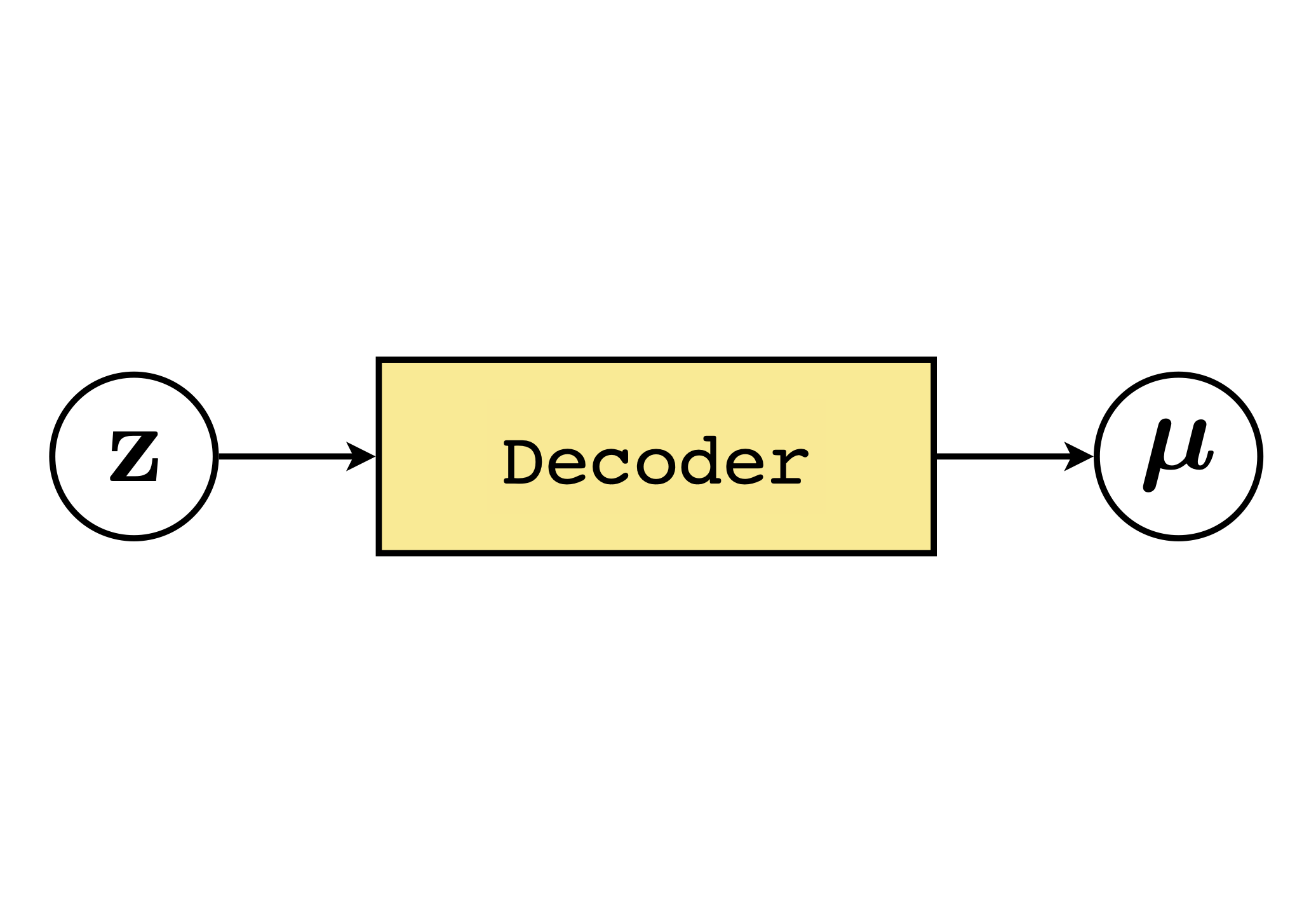}
        \caption{VAE}
        \label{fig:vae}
    \end{subfigure}
    \begin{subfigure}[b]{0.31\linewidth}
        \centering
        \includegraphics[width=\textwidth]{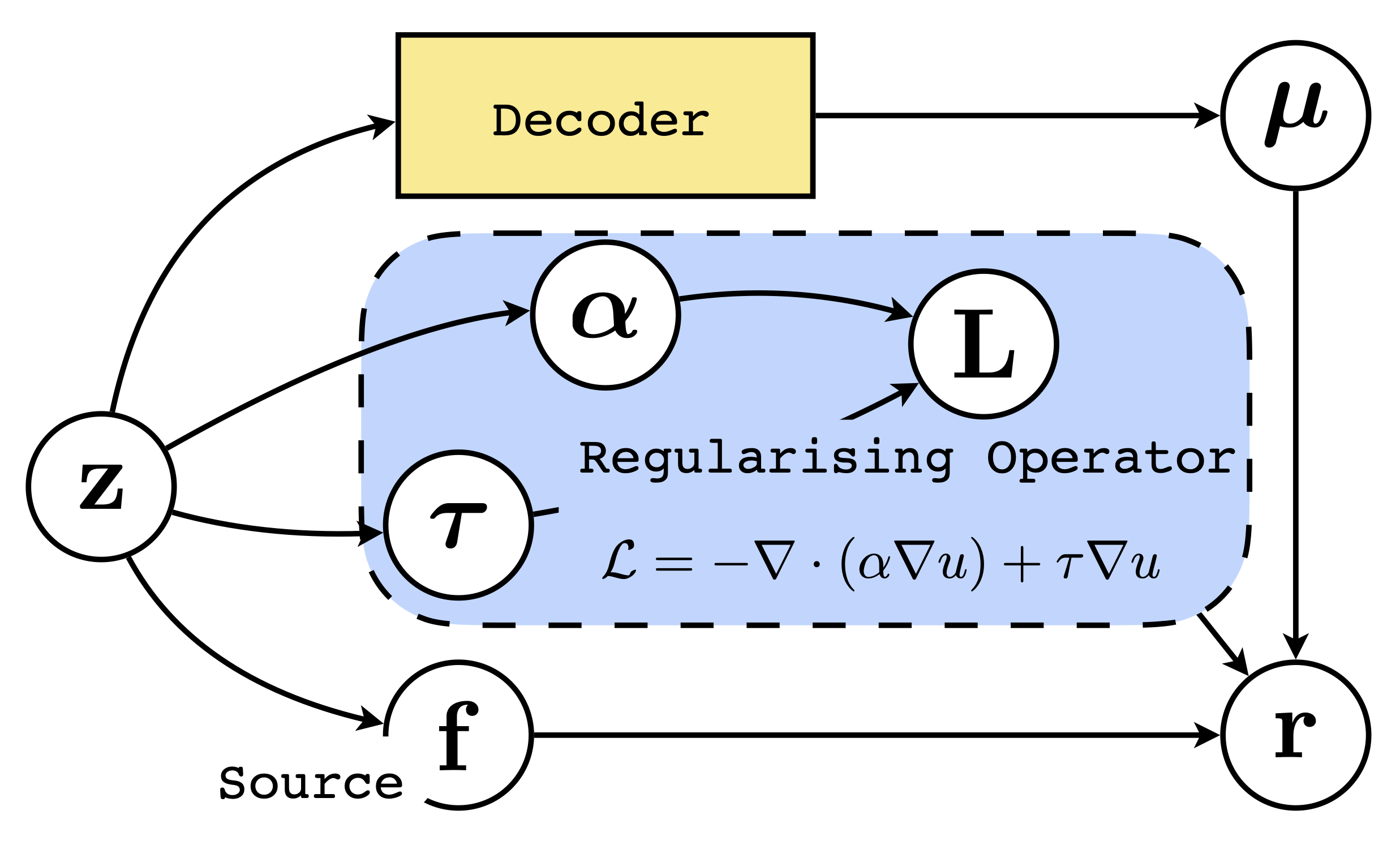}
        \caption{\texttt{VAE-DT}}
        \label{fig:vae_dt}
    \end{subfigure}
    \hspace{.1cm}
    \begin{subfigure}[b]{0.31\linewidth}
        \centering
        \includegraphics[width=\textwidth]{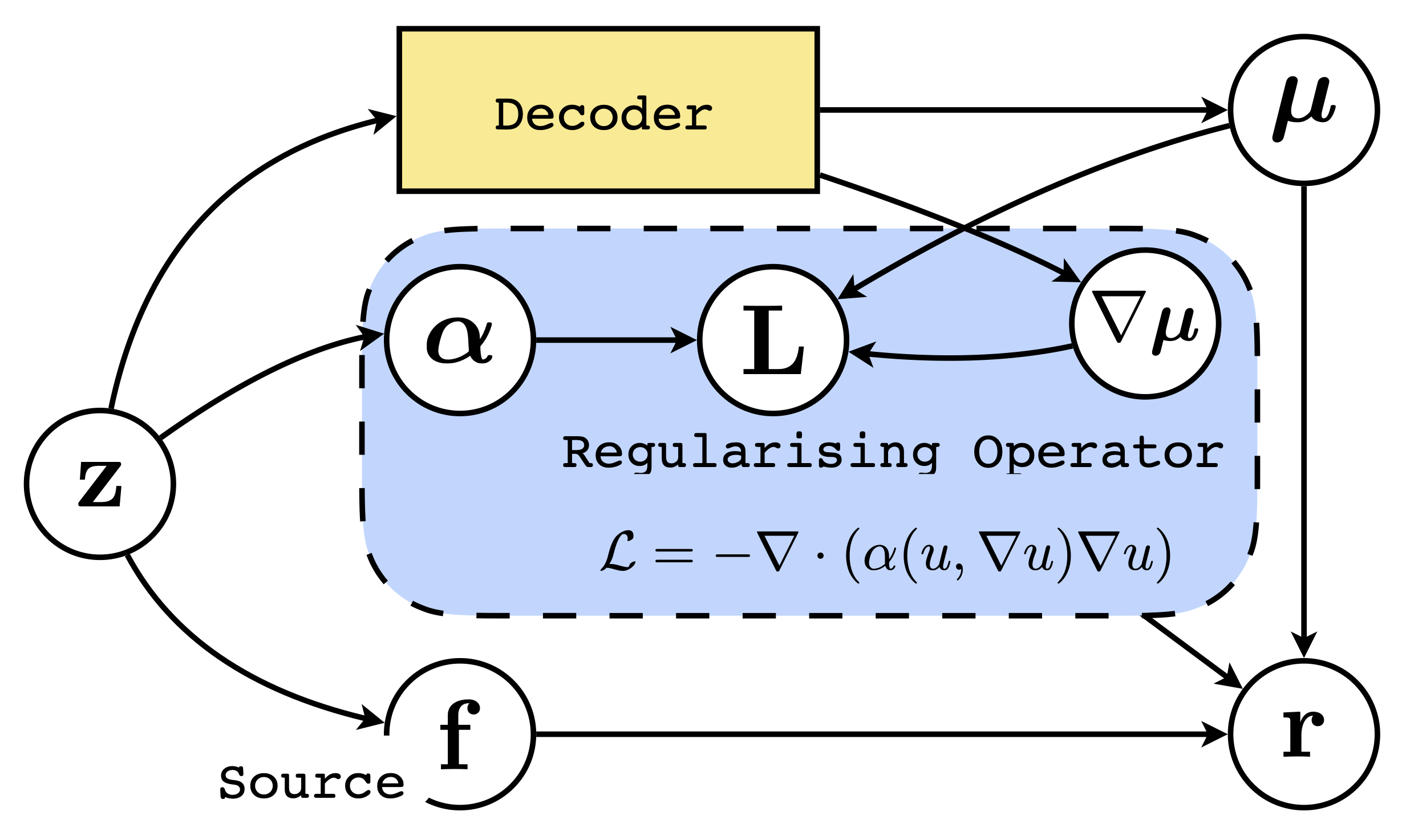}
        \caption{\texttt{VAE-NLD}}
        \label{fig:vae_nld}
    \end{subfigure}
    \caption{PDE regularised decoder networks for the generative models in the 
    Jura experiment. (a) presents the base VAE model, (b) and (c) augment the decoder network
    with a regularising PDE operator and an input source, 
    for full details of the architechtures
    used see Sec. 4 of the supplement.}
    \label{fig:my_label}
\end{figure}

Our aim in this section is threefold; to demonstrate the accuracy of
our method compared to ground-truth, to quantify computational efficiency
and finally to demonstrate applicability to real world problems. All experiments were run on a 2.9 GHz i9 processor with 
2400MHz RAM. See the supplement for discussion on the architectures and training setup used for each experiment.

\subsubsection*{Mini-patching accuracy and efficiency} We 
perform the experiment depicted in 
Fig. \ref{fig:pde_as_smoothers}, simulating 
from \eqref{eq:transport} 
with constant source on increasingly fine meshes. 
To assess our PDE constrained
VI method we implement it with varying tapering chosen so that
each sub-mesh had on average $Q_{\text{int}} \in \{32, 64, 128\}$
nodes, the specifications are labelled as $\text{CVI}_{Q_{\text{int}}}$. 
These are compared to the benchmark Hamiltonian Monte Carlo 
(HMC) \citep{duane}. 
We learn the posterior of the transport vector field, 
and compare the mean absolute error (MAE) on validation data using samples from the learned models.

\begin{figure}[!htb]
    \centering
     \begin{subfigure}[b]{.45\linewidth}
         \includegraphics[width=\textwidth]{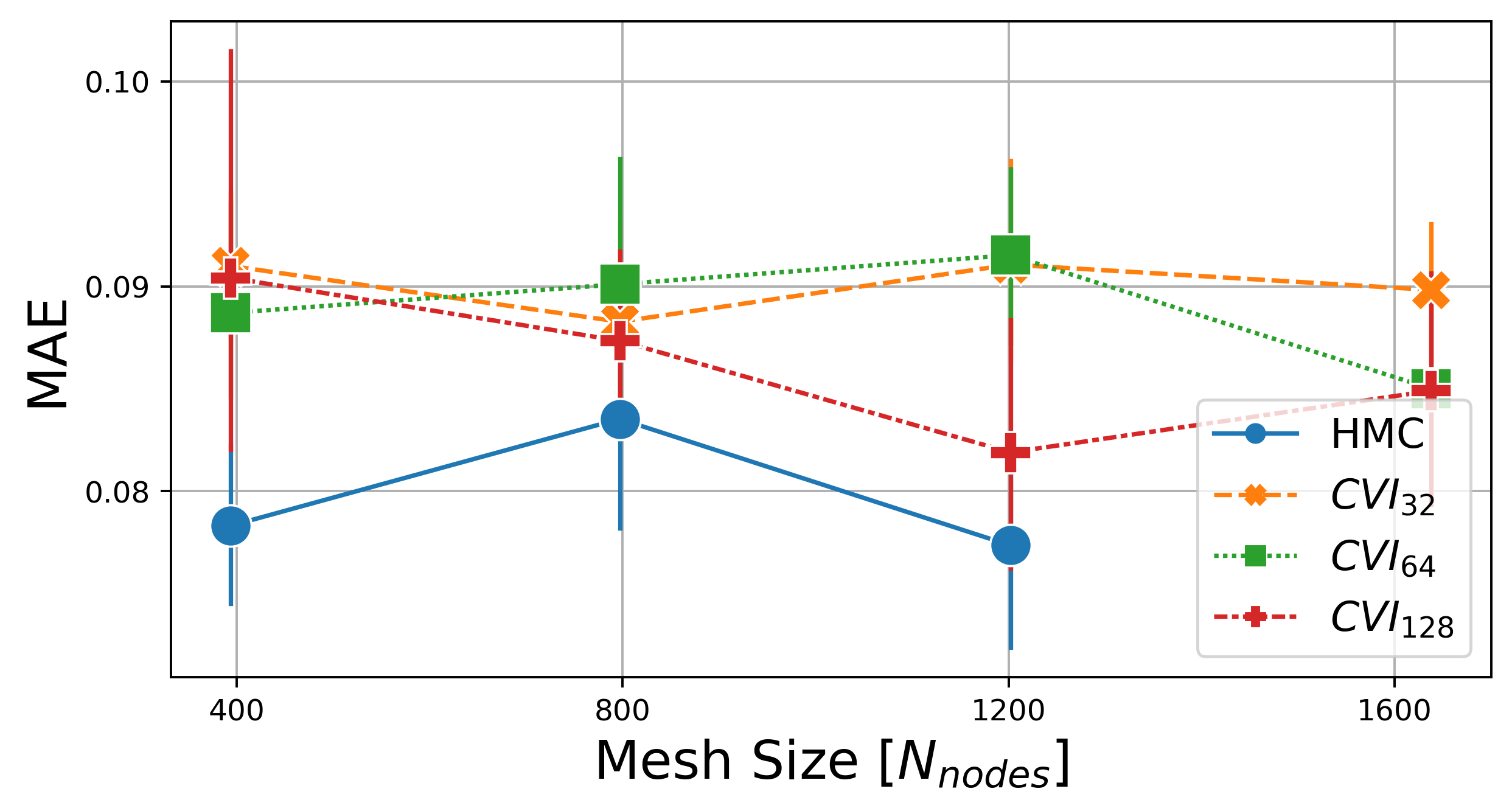}
         \caption{Accuracy}
         \label{fig:accuracy_exper}
     \end{subfigure}\hfill
     \begin{subfigure}[b]{.45\linewidth}
         \includegraphics[width=\textwidth]{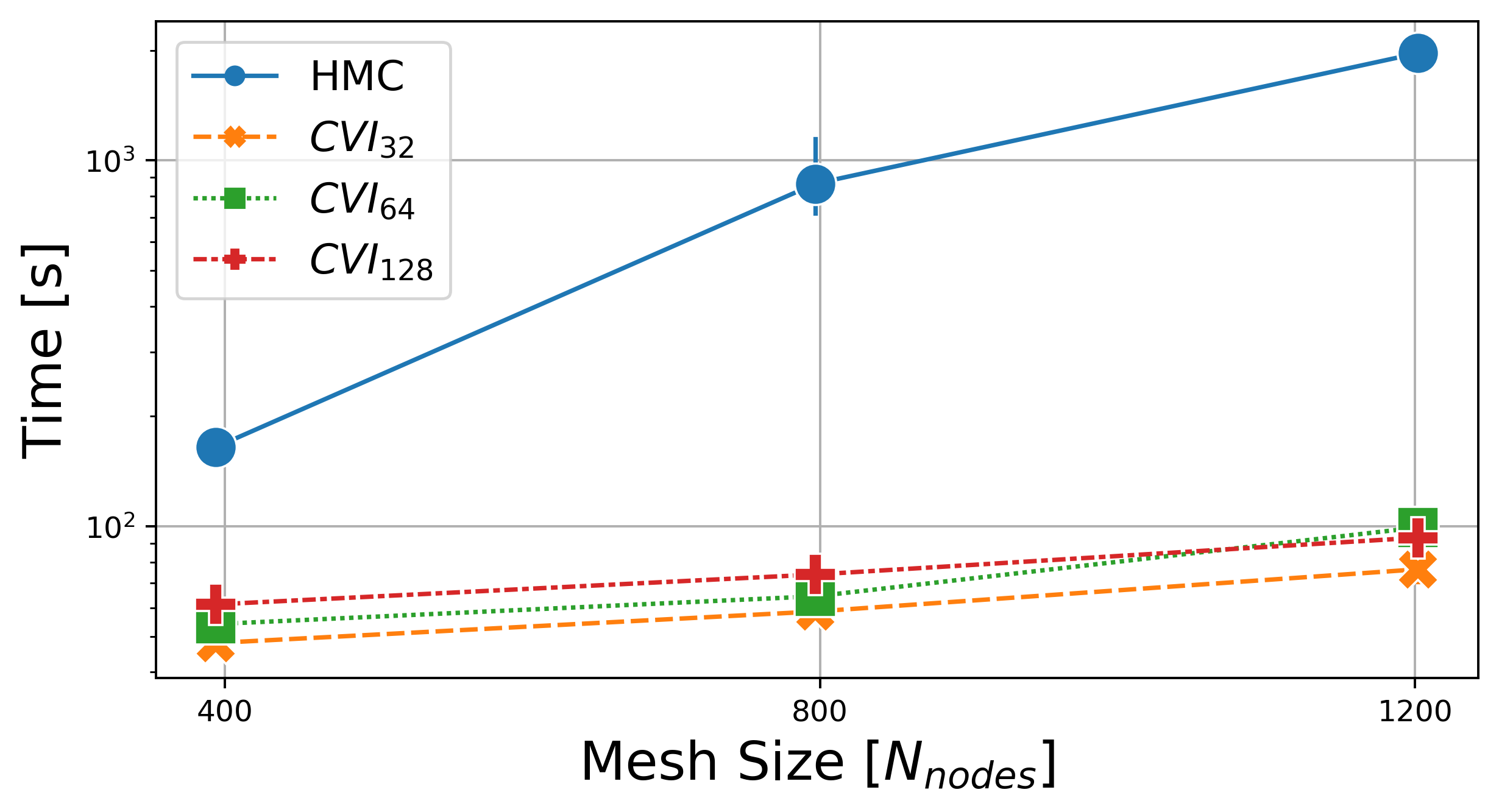}
         \caption{Total Run-time}
         \label{fig:time_compar}
     \end{subfigure}
    \caption{
    Effect of increasing mesh resolution on accuracy
    and total run-time of our accelerated local
    approximation compared to ground truth HMC 
    on 10 rpelicates of the transport problem. 
    Results of HMC on the finest mesh are not 
    shown because time taken exceeded 
    computational budget.}
\end{figure}

Results are presented in Fig. \ref{fig:accuracy_exper} for all 
methods after a total of 10000 training iterations, as expected 
the HMC methods 
shows the best absolute performance. However, even at 
the smallest patch size we are able to match performance within error bounds, and
at a substantially reduced cost. Fig. \ref{fig:time_compar}
reports total run-time on a log-scale where we observe the
$\mathcal{O}(\Nnodes^3)$ scaling of the \texttt{Solve} in the HMC method,
conversely our regulariser scales as $\mathcal{O}(N_{\text{mini-patch}}^2)$
where $N_{\text{mini-patch}}$ is the maximum number
of nodes in a mini-patch, allowing our method to be applied
even on very dense meshes.

\subsubsection*{Heavy metal contamination, Swiss Jura} Diffusion 
and transport through topsoil and 
waterways can lead to wide dispersal of contaminants.
Since these physical influences 
are common to each contaminant we can use the 
presence of \emph{secondary metals}
to infer the concentration of a \emph{primary metal},
here cadmium, copper, lead and cobolt, and this analysis
is undertaken by \citep{goovaerts} via kriging. 

To demonstrate our method as a plug-in enhancement
to a generic model we shall consider a VAE
\citep{kingma, rezende14}, with generic decoder
Fig. \ref{fig:vae}, and then add our mechanism-based 
regularisation. First \texttt{VAE-DT}, 
the PDE \eqref{eq:transport} with diffusion and transport which will
have the generative structure in Fig. \ref{fig:vae_dt},
and then \texttt{VAE-NLD}, a nonlinear model with diffusion operator 
$\mathcal{L}[\bz] = -\nabla \cdot (a(\bz, u, \nabla u) \nabla u)$ shown in Fig. \ref{fig:vae_nld}.
Allowing the diffusion coefficient to be a function of the field
variable and the gradient has been shown
\citep{RUDIN1992259} to lead to stronger feature preservation. 
We further compare our model to the 
GP diffusion kernel model of \citep{alvarez_2009}, a linear
PDE with constant diffusion and no transport. Results are displayed in 
Table \ref{tab:jura_results}; these demonstrate that
the VAE was consistently outperformed,
but that once our physically informed regularisation was
included it was able to outperform or match accross
all settings. We stress that 
the base architecture was constant
for each VAE, indicating the ability of the 
method we introduce in this paper to act as a drop-in 
enhancement to existing architectures.

\begin{table}[ht]
\caption{MAE from ten repetitions of prediced heavy metal 
concentration on the Jura dataset. GPDK is the diffusion kernel \citep{alvarez_2009}. \texttt{VAE-DT} and 
\texttt{VAE-NLD} extend VAE with our PDE 
regularisation. Results
in bold indicate significance under a Wilcoxon test comparing each VAE model to the
GPDK}
\label{tab:jura_results}
\vskip 0.15in
\begin{center}
\begin{small}
\begin{sc}
\begin{tabular}{lcccc}
\toprule
& GPDK \citep{alvarez_2009} 
& VAE \citep{kingma, rezende14} & \texttt{VAE-DT} 
& \texttt{VAE-NLD} \\
\midrule
Cd & $\boldsymbol{0.451 \pm 0.013}$ & $0.569 \pm 0.115$ & $0.478 \pm 0.048$ & $0.549 \pm 0.173$ \\
Cu & $\boldsymbol{7.168 \pm 0.347}$ & $7.752 \pm 0.341$ & $7.218 \pm 0.315$ & $7.625 \pm 0.201$ \\
Pb & $10.101 \pm 0.284$ & $15.69 \pm 0.294$ & $10.058 \pm 0.297$ & $\boldsymbol{9.722 \pm 0.251}$ \\
Co & $1.755 \pm 0.090$ & $1.820 \pm 0.102$ & $1.801 \pm 0.173$ &$\boldsymbol{1.692 \pm 0.083}$ \\
\bottomrule
\end{tabular}
\end{sc}
\end{small}
\end{center}
\vskip -0.1in
\end{table}

\subsubsection*{Eastern Snake River Plain Aquifer, Idaho}
An aquifer is an underground layer of permeable rock, from 
which groundwater can be extracted. Geological properties 
govern how water permeates, 
and local hydrological features act as additional inputs 
to the system. To capture these physical processes we 
consider two different PDE specifications, each having a 
latent GP source function. The first, \texttt{GP-D}, 
possesses an inhomogenous log-GP diffusion coefficient 
capturing spatially varying 
diffusion, the second, \texttt{GP-DT}, is augmented 
with a spatially homogeneous transport field. 
\begin{wrapfigure}{r}{0.5\linewidth}
    \includegraphics[width=7cm]{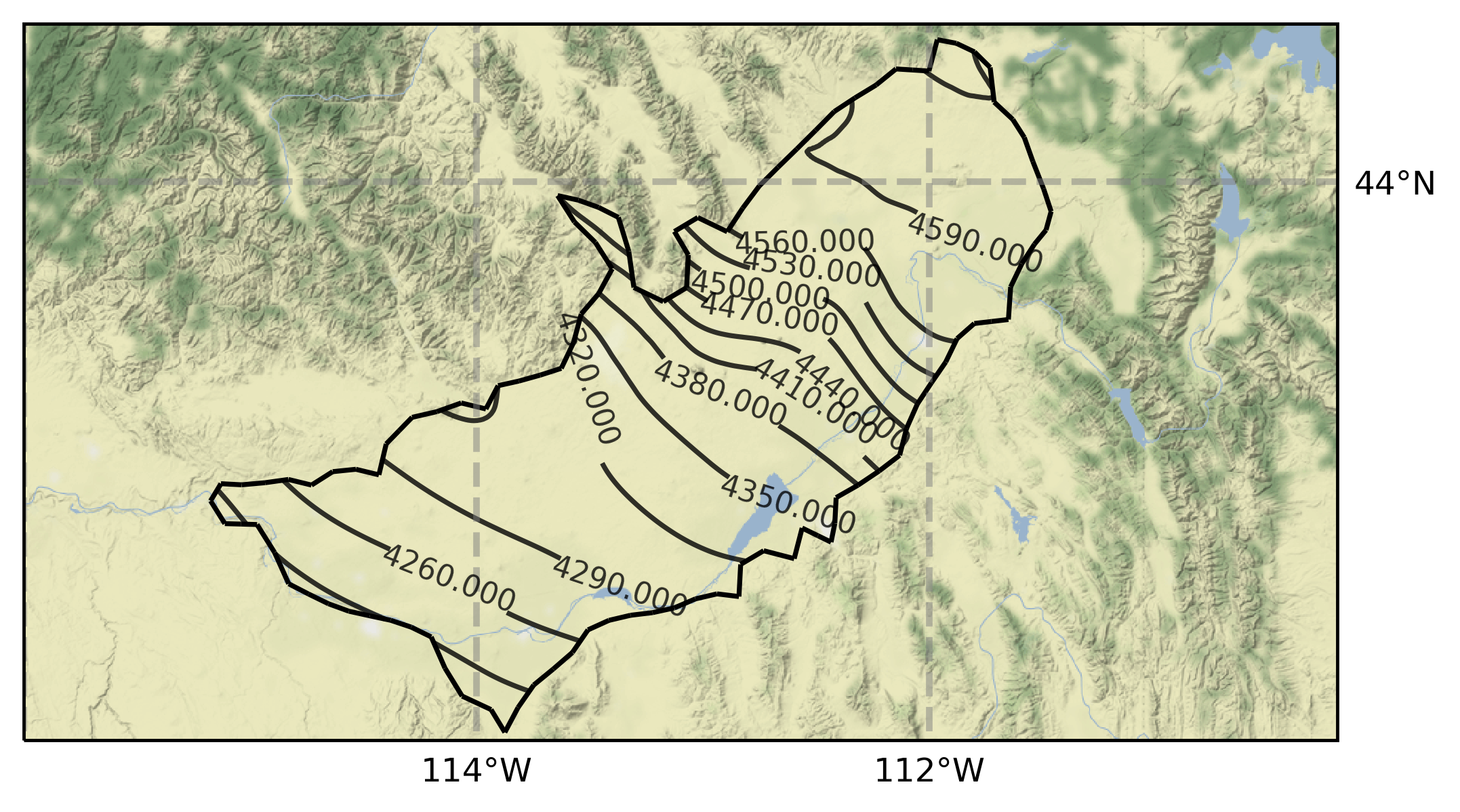}
    \caption{Mean predicted depth (in feet) to the
    groundwater level from our \texttt{GP-DT}
    }
    \label{fig:aquifer}
\end{wrapfigure}
The resulting generative structure is similar to that of Fig.
\ref{fig:vae_dt}, see Sec. 4 of the supplement for the full
presentation of the model. We train on levels from 202 wells over the 
period 1960--1980 reported in \citep{ackerman}, and predict on 
a further set of 242 measurements from 1980--2000
\footnote{Well measurements are available from 
\url{https://water.usgs.gov/ogw/networks.html}}. 

Our PDE influenced specifications are compared with baseline predictions from 
models encoding no physical structure; namely a GP with
Matern kernel and, to allow more complex data-driven patterns, a 
2-layer Deep GP (DGP) \citep{damianou_2013}. Results are displayed in
Table \ref{tab:aquifer_results} where we find that the PDE 
regulated models, embodying the richest physical structure
outperform the mechanistically simpler alternatives. The 
predicted groundwater level 
is displayed in Fig. \ref{fig:aquifer}, agreeing 
with the output of numerical work presented in \citep{ackerman}. 
To view additional figures, including the estimated parameters, see
the supplementary material.

\begin{table}[ht]
\caption{Validation error for the aquifer data on well measurements 
from 1980--2000. \texttt{GP-D} is our model with GP source and 
log-GP diffusion coefficient, 
\texttt{GP-DT} is further augmented with a transport vector field. We report
mean absolute and mean squared error ($\pm2$ standard deviations).}
\label{tab:aquifer_results}
\vskip 0.15in
\begin{center}
\begin{small}
\begin{sc}
\begin{tabular}{lcccc}
\toprule
& GP & DGP \citep{damianou_2013} & \texttt{GP-D} & \texttt{GP-DT}\\
\midrule 
MAE & $0.224 \pm 0.041$ & $0.193 \pm 0.052$ 
& $0.160 \pm 0.012$ & $\boldsymbol{0.158 \pm 0.001}$ \\
MSE & $0.107 \pm 0.050$ & 
$0.081 \pm 0.011$
& $0.044 \pm 0.001$ & $\boldsymbol{0.039 \pm 0.001}$ \\
\bottomrule
\end{tabular}
\end{sc}
\end{small}
\end{center}
\vskip -0.1in
\end{table}

\section{Discussion}
\label{sec:discussion}
We have considered the problem of accelerating
the BIP by constructing VI approximations which respect
the generative structure of the mechanistic model. By
taking an optimisation-centric view beginning in the dual
space we are able to soften the hard constraints of the 
original problem and implement a method which is able 
to achieve increased computational efficiency, 
without sacrificing accuracy. Our experiments demonstrated 
that our approach can be
used both to accelerate the classical BIP and as a 
drop-in enhancement to more general methods. Beyond 
this we have introduced a framework which allows for 
the uncertainty in the governing dynamics, enabling 
what prior physical knowledge exists to be easily 
combined with flexible ML methods allowing one to 
leverage the full power of these methods, 
without ignoring the wealth of scientific knowledge 
already available.

\subsection*{Broader Impact}

Our work both accelerates the classical inverse
problem, and offers improved 
uncertainty quantification (UQ) in the presence or prior, 
but possibly incomplete, physical
knowledge, arguably the most realistic knowledge
state in real-world applications. By designing
our method as a plug-in enhancement, and by virtue of
the predominance of the FEM in industry
\citep{gupta}, we ensure that our method will be more 
immediately familiar to engineers than model free methods.
Thereby allowing for improved uptake of
ML techniques in industries which have been slower
to adopt statistical methods compared to newer technological 
sectors. Improved UQ in these industries is vital for
reasons including the specification
of warranties and providing failure prediction and prevention, furthermore 
our application to contaminants and water-level modelling is of
immediate importance to public policy in drought 
afflicted regions, or ensuring safe drinking water
in industrialised regions. Given the high 
potential human cost in all of the above mentioned use cases it 
is vital that the predictions on which warranty or policy decisions are based
be transparent and accountable for. Transparency and interpretability can be particularly problematic
for general DL methods, however
by including a mechanistic component inside the
decoder architecture we allow for a level of 
structure-based interpretation that would not be possible with a
purely data-driven deep generative method.

Furthermore, our mini-patching idea invites future 
study into how local information can be better used 
to regularise global mechanistic models. As edge based 
computing continues to grow in importance integrating local information into global models will become increasingly fundamental. We have integrated well-level readings into a large 
scale hydrological model, but such an approach would apply equally to using phones 
and wearable technology as local pressure/temperature sensors as inputs into large-scale climate models. Rightly a great deal of importance is placed
on the privacy of an individuals location data, and it will therefore be important
that future work on this front proceeds in a way that both respects privacy, but also fully utilises the potential of local models to inform a global mechanism which our work has begun to develop.

\bibliographystyle{plainnat}
\bibliography{refs.bib}

\appendix
\section*{Appendices}
\addcontentsline{toc}{section}{Appendices}
\renewcommand{\thesubsection}{\Alph{subsection}}





\section{Additional details of the FEM}
This paper has aimed to combine several areas, most
notably the solution of PDEs by the finite element method,
and the use of deep probabilistic generative models in
machine learning. We therefore provide some additional
information regarding the FEM method, mostly concerning
notational points, that was not
included in the main body, however for a 
more comprehensive review see 
\citep{reddy, brenner}.

\begin{figure}[h]
    \centering
    \begin{subfigure}[b]{.31\linewidth}
    \centering
        \includegraphics[height=3cm]{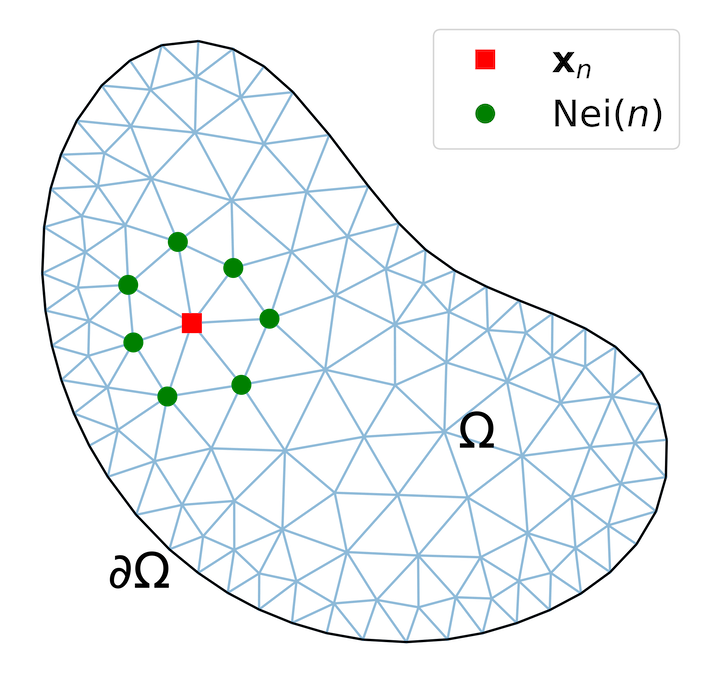}
        \caption{FEM Mesh}
        \label{fig:fem_explan_a}
    \end{subfigure}
    \hfill
    \begin{subfigure}[b]{.31\linewidth}
    \centering
        \includegraphics[height=3cm]{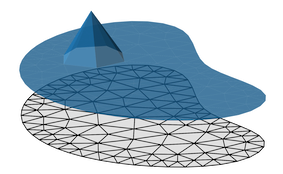}
        \caption{A nodal basis function}
        \label{fig:nodal_basis_func}
    \end{subfigure}
    \hfill
    \begin{subfigure}[b]{.31\linewidth}
    \centering
        \includegraphics[height=3cm]{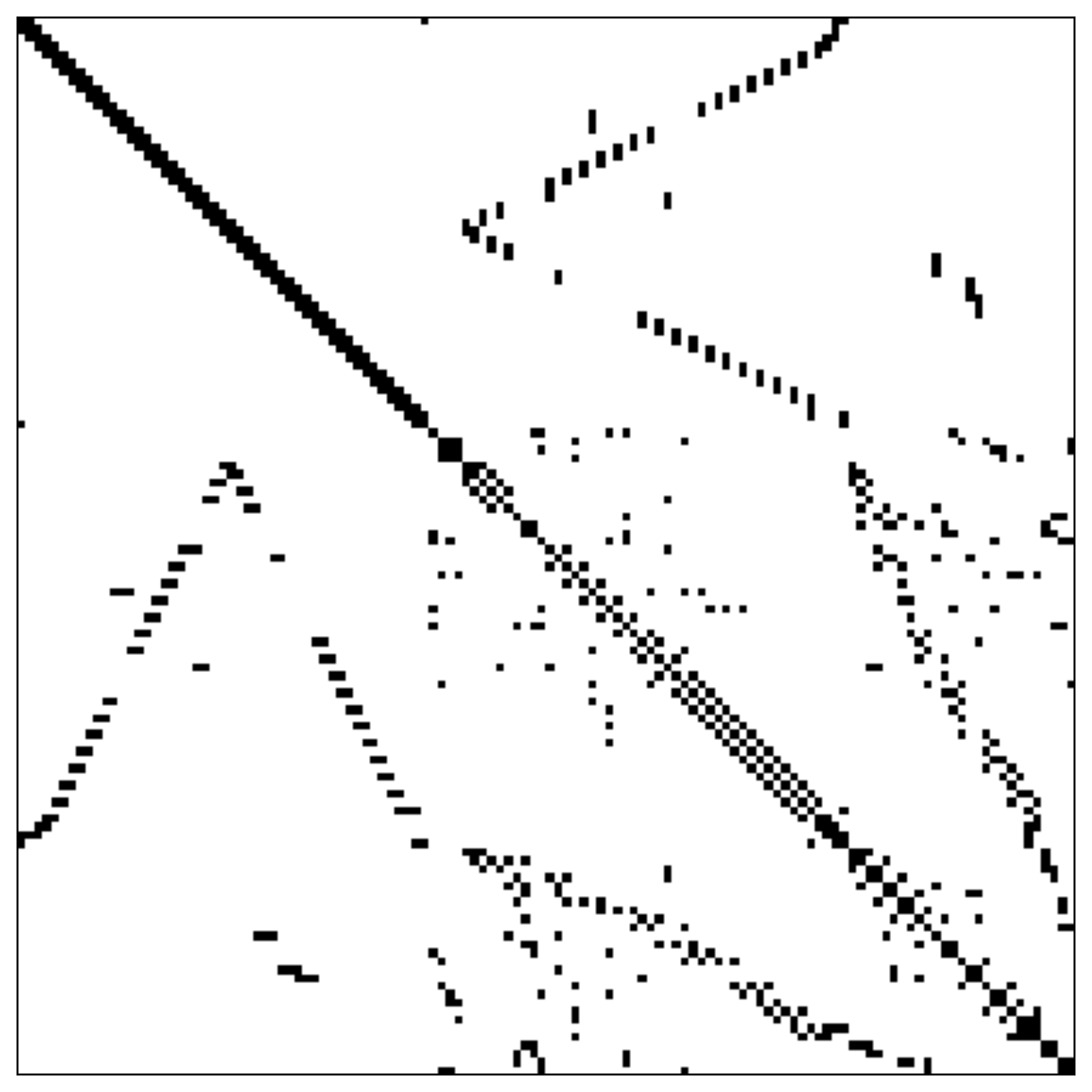}
        \caption{Stiffness matrix sparsity}
        \label{fig:stiffness_matrix_sparsity}
    \end{subfigure}
    \caption{(a) Decomposition of a spatial domain $\Omega$ into a FEM mesh, 
    $\Omega = \cup_{k \in \mathbb{T}}\Delta_k$.
    (b) An example nodal basis function.
    (c) Sparsity pattern of the 
    stiffness matrix after assembly using nodal basis functions, upper-left 
    block corresponds to boundary nodes which have lower connectivity.}
\end{figure}

When necessary we shall denote a node in the mesh using bar notation, 
and the complete set of nodes by $\{ \bar{\bx}_j \}_{j=1}^{\Nnodes}$.
A FEM mesh naturally applies an adjacency structure to this node
set with node $j \in \mathrm{nei}_{\Delta}(i)$ if there is
an edge of the mesh between these two nodes, this is
depicted in Fig. \ref{fig:fem_explan_a}. 
We use the notation $\mathrm{nei}_{\Delta}$ to distinguish this adjacency structure arising from the mesh elements with the adjacency
structure introduced in Sec. \ref{sec:fwd_model} from the
tapering function.

The basis function used in the FEM is typically a
\emph{nodal} basis function, which is characterised by
the fact that
\begin{align*}
    \phi_i(\bar{\bx}_j) = \delta_{ij}
\end{align*}
this leads to the incredible sparsity of the FEM
construction. Indeed the functions $\phi_i$ now
only have support on the set of elements with 
$\bar{\bx}_i$ as a vertex, this is displayed in 
Fig. \ref{fig:nodal_basis_func}. This leads to
a much smaller quadrature in the variational problem
\eqref{eq:finite_var_problem} and the assemblys of
\eqref{eq:stiffness_and_load_vector} or
\eqref{eq:transport_assembly}. This is seen in 
Fig. \ref{fig:stiffness_matrix_sparsity} where
plot the sparsity patern of a typical 
stiffness matrix $\mathbf{A}$.

We noted in Sec. \ref{sec:review} that solving the PDE problem
then involves solving for the coefficient $\boldsymbol{\xi}^u$
in equations such as
\begin{align*}
    \mathbf{A} \boldsymbol{\xi}^u = \bff,
\end{align*}
and this defines a function $u(\bx) = \sum_{m=1}^{M} 
(\boldsymbol{\xi}^u)_m \phi_m(\bx)$ in $V$. For notational
convenience we shall also refer to the vector $\boldsymbol{\xi}^u$
as a ``function'', with the understanding that when we do so
we are actually referring to the just described expansion. When
doing so we shall use the more direct notation $\mathbf{u}$ for
this finite dimensional representation of the function $u(\bx)$.

\section{Additional details for Section \ref{sec:struct_preserving_VI}}
\subsection{Dual-space perturbation}
In this section we provide additional details concerning
our dual-space pertubation introduced in Sec. 
\ref{sec:relaxing_the_vi_prob}, doing so requires some
familiarity with the concept of a Sobolev space, 
a particular instance of a reproducing kernel 
Hilbert space (RKHS) and its dual space, for an
overview of this functional analytic material in
a PDE setting we recommend \citep{brezis}.

While one often encounters the Laplace operator
in the form presented in \eqref{eq:poisson} as a
differential operator in
the strong sense taking a twice-differentiable
function and outputting a new function, we can
also consider an operator 
$-\Delta \; : \; H_0^1 \rightarrow H^{-1}$ from
the Sobolev space $H_{0}^{1}$ to its dual space
$H^{-1}$. This operator, which we refer to as
the \emph{(negative) Laplacian}, takes a function
$u \in H_0^{1}$ and maps it to the continous
linear function $-\Delta[u]$ say, which acts on
functions $v \in H_0^1$ by
\begin{align}
    -\Delta[u] v = \int_{\Omega}
    \nabla u(\bx) \cdot \nabla v(\bx) d\bx.
\end{align}
This operator has an inverse, the \emph{inverse negative
Laplacian}, or more simply just the 
\emph{inverse Laplacian} and denoted $(-\Delta)^{-1}$.
The inverse Laplacian is then an operator
$(-\Delta)^{-1} \; : \; H^{-1} \rightarrow H_{0}^{1},$
moreover this operator defines a Riesz-Fr\'echet 
isomorphism between the Sobolev space $H_0^1$ of
functions with weak-derivative \citep{brezis} 
and its dual-space \citep{brezis, ito_kunisch_1990} 
and therefore
\begin{align}
    \| \varphi \|_{H^{-1}} = \| (-\Delta)^{-1}\varphi \|_{H^1_0},
    \label{supp:eq:rf_iso}
\end{align}
where $\| \cdot \|_{H_{0}^{1}}$ is the Sobolev norm induced by the inner
product
\begin{align}
        \langle u, v \rangle_{H^1_0} = 
        \int_{\Omega} \nabla u(\bx) \cdot \nabla v(\bx) d\bx,
\end{align}
and applying the divergence theorem we also have the identity
\begin{align}
    \langle u, v \rangle_{H^1_0}
    = - \int_\Omega \Delta u(\bx) v(\bx) d\bx 
    \stackrel{\Delta}{=} \langle -\Delta u, v \rangle_{L^{2}(\Omega)} \label{supp:eq:sob_l2}
\end{align}
this change of sign after applying the divergence theorem
now justifies the slight inconvenience of defining everything 
earlier in terms of the negative Laplacian operator.

Taken together the results above give us the chain of identities used in Section
\ref{sec:relaxing_the_vi_prob}, in particular we 
first note that the weak form defines an element
of the dual space, which will act on a function $v$ by
\begin{align}
    \varphi_{u, \bz} \; : \; v \mapsto \int_{\Omega} \mathcal{L}u(\bx) v(\bx) d\bx 
    - \int_{\Omega} f(\bx) v(\bx) d\bx
\end{align}
which we use the FEM to approximate as the element
\begin{align}
    \varphi_{u, \bz} \approx 
    \mathbf{L}[\bz]\bu - \mathbf{f}
    \label{supp:eq:discrete_form}
\end{align}
where $\bu$ is the finite vector of coefficients
obtained after projecting the function $u(\bx)$ onto the
finite dimensional subspace
$V_M \subset H_0^1$. This is what we mean by the
discretised version of the weak form \eqref{supp:eq:discrete_form}, and
the discrete form can be interpreted as giving us the 
image of the basis vectors $\hat{\phi}$, that is
\begin{align}
    \varphi_{u, \bz} (\hat{\phi}_j) \approx 
    \left(\mathbf{L}[\bz]\bu - \mathbf{f}\right)_j.
\end{align}

Now to construct our approximate measure we shall
start from the infinite-dimensional picture in the
dual space, and then discretise at the end and so we
briefly set aside the discrete version of the weak
form operator just introduced. Then beginning 
from the Gaussian measure\footnote{That this is
indeed a measure is mostly directly seen from the
just described isomorphism between the
RKHS $H_0^1$ and $H^{-1}$ with the norm 
induced by the Sobolev inner product.}
\begin{align}
    \exp\left\{ -\frac{1}{2\epsilon^2}
    \| \varphi_{u, \bz} \|_{H^{-1}}
    \right\}
\end{align}
we apply the isomorphism \eqref{supp:eq:rf_iso} and
then the definition \eqref{supp:eq:sob_l2} to get
\begin{align}
    \exp\left\{ -\frac{1}{2\epsilon^2}
    \| \varphi_{u, \bz} \|_{H^{-1}}
    \right\} &=
        \exp\left\{ -\frac{1}{2\epsilon^2}
    \| (-\Delta)^{-1}\varphi_{u, \bz} \|_{H_{0}^1}
    \right\} \notag \\
    &=  \exp\left\{ -\frac{1}{2\epsilon^2}
    \langle (-\Delta) (-\Delta)^{-1}
    \varphi_{u, \bz}, 
    (-\Delta)^{-1} \varphi_{u, \bz} 
    \rangle_{L^2(\Omega)}
    \right\} \notag \\
    &=  \exp\left\{ -\frac{1}{2\epsilon^2}
    \langle
    \varphi_{u, \bz}, 
    (-\Delta)^{-1} \varphi_{u, \bz} 
    \rangle_{L^2(\Omega)}
    \right\} 
\end{align}
Now we project all of the infinite-dimensional elements
onto their finite dimensional representations so we replace
$\varphi_{u, \bz}$ with \eqref{supp:eq:discrete_form} and $(-\Delta)^{-1}$ by
the inverse stiffness matrix $\mathbf{A}^{-1}$ to give the
finite-dimensional Gaussian measure
\begin{align}
    \exp\left\{ -\frac{1}{2\epsilon^2}
    (\mathbf{L}[\bz]\bu - \bff)^{\top}
    \mathbf{A}^{-1}
    (\mathbf{L}[\bz]\bu - \bff)
    \right\} 
\end{align}
which is exactly our approximating measure for the relaxed weak form problem.

In Sec. \ref{sec:struct_preserving_VI} and Sec. \ref{sec:fwd_model}
we repeatedly penalise our model by the expression
\begin{align*}
    \| \mathbf{L}[\bz]\mu(\bz) - \bff\|_{\bA},
\end{align*}
the derivations above allow us to understand this intuitively
as measuring the deviation of the
dual-space element $\varphi_{\mu, \bz}$ obtained using the 
forward-surrogate from zero, which would be the value of the
element $\varphi_{G[\bz], \bz}$ where we recall from 
Sec. \ref{sec:review} that $G$ is the forward map, i.e. the
implicit solution of the true PDE. We are then measuring the
scale of these deviations under a Gaussian measure centered on
zero with scale parameter $\epsilon$, and we view
the measured elements $\varphi_{\mu, \bz}$ which no 
longer satisfy the PDE constraint exactly as 
``approximate mechanisms'' as remarked at the end of 
Sec. \ref{sec:struct_preserving_VI}. 

\subsection{The optimisation problem}
In general the forward surrogate $\mu$ will depend on the 
spatial coordinate, however to improve presentation we 
suppress this dependence in the following and simply 
write $\mu(\bz)$ to denote the dependence
of this variable on the latent processes.

Recall from Section \ref{sec:struct_preserving_VI} 
that we choose to parameterise the conditional 
variational factor of the forward model as
the Gaussian
\begin{align}
    q(\bu \mid \bz) = \mathcal{N}(
    \bu \mid \mu(\bz), \epsilon^2(\mathbf{L}[\bz]\mathbf{A}^{-1} \mathbf{L}[\bz])^{-1}),
\end{align}
also recall that our target objective function is given by
\begin{align}
    F_{\epsilon} =
    \mathbb{E}_{\bu \sim q(\bu, \bz)}\left[
    - \log p(\by \mid \bu)
    \right]
    + \KL( q(\bu, \bz) \| p_{\epsilon}(\bu \mid \bz)p(\bz) ).
\end{align}
Then after applying the ``chain rule for divergences'' to expand 
$\KL(q(\bu, \bz) \| p_{\epsilon}(\bu \mid \bz)p(\bz))$ we seek to 
minimize the following variational lower bound
\begin{align}
    &\mathbb{E}_{\bz \sim q(\bz)} 
    \left[ \mathbb{E}_{\bu \sim q(\bu \mid \bz)} \left[ 
    -\log p(\by \mid \bu) \right] \right] \notag \\
    &\qquad + \KL(q(\bz) \| p(\bz) ) \notag \\ 
    &\qquad + \mathbb{E}_{\bz \sim q(\bz)}\left[ 
    \KL(q(\bu \mid\bz) \| p_{\epsilon}(\bu \mid \bz) )
    \right]
\end{align}
Focusing on the final term, and using the fact that 
the conditional covariance matrices in 
$q(\cdot \mid \bz)$ and $p_{\epsilon}(\cdot \mid \bz)$ 
match, we have
\begin{align}
&\mathbb{E}_{\bz \sim q(\bz)}\left[ 
    \KL(q(\bu \mid\bz) \| p_{\epsilon}(\bu \mid \bz) )
    \right] \notag \\
&\qquad = \frac{1}{2\epsilon^2}\mathbb{E}_{\bz \sim q(\bz)}\left[ 
    (\mu(\bz) - \mathbf{L}[\bz]^{-1}\mathbf{f})^{\top}
    \mathbf{L}[\bz]\mathbf{A}^{-1}\mathbf{L}[\bz]
    (\mu(\bz) - \mathbf{L}[\bz]^{-1}\mathbf{f})^{\top}
    \right] \notag \\
    &\qquad = \frac{1}{2\epsilon^2} \mathbb{E}_{z \sim q(\bz)}\left[ 
    \|\mathbf{L}[\bz]\mu(\bz) - \mathbf{f} \|_{\mathbf{A}}^2
    \right].
\end{align}
From which we get the expression 
\eqref{eq:cvi_sub_prob} in the main body of the text.

If we now let $\{ \bz \}_{i=1}^{M}$ be a collection of 
independent samples from $q(\bz)$, then 

\begin{align}
    \frac{1}{2\epsilon^2}\mathbb{E}_{z \sim q(\bz)}\left[ 
    \|\mathbf{L}[\bz]\mu(\bz) - \mathbf{f} \|_{\mathbf{A}}^2
    \right]
    \approx \frac{1}{M} \sum_{i=1}^M
    \| \mathbf{L}[\bz_i]\mu(\bz_i) - \mathbf{f} \|_{\mathbf{A}}^2
\end{align}

The full approximate objective function after applying a Monte-Carlo approximation
to the objective function is therefore given by

\begin{align}
    F_{\epsilon} \approx \mathbb{E}_{\bu \sim q(\bu, \bz)}
    \left[- \log p(\by \mid \bu) \right] 
    + \KL(q(\bz) \| p(\bz))
    + \frac{1}{2\epsilon^2 M} 
    \sum_{i=1}^M
    \| \mathbf{L}[\bz_i]\mu(\bz_i) - \mathbf{f} \|_{\mathbf{A}}^2.
    \label{supp:eq:Feps}
\end{align}

Defining the variable 
$\mathbf{r}_i \stackrel{\Delta}{=}\mathbf{L}[\bz_i]\mu(\bz_i) - \mathbf{f}$,
then we recognise \eqref{supp:eq:Feps} as the quadratic penalty form of the 
following objective function
\begin{subequations}
\begin{empheq}{alignat=2}
    &\argmin_{q(\bz) \in \mathcal{Q}, \mu \in V}
    \mathbb{E}_{\bu \sim q(\bu, \bz)}
    \left[- \log p(\by \mid \bu) \right] 
    + \KL(q(\bz) \| p(\bz)) \\
    &\mbox{subject to} \quad \| \mathbf{r}_i \|_{\bA} = 0, \qquad i=1,\ldots, M,
\end{empheq}
\end{subequations}
with $M$ constraints. Alternatively, and because the sample was arbitrary, 
we conclude that
\begin{subequations}
\begin{empheq}{alignat=2}
    &\argmin_{q(\bz) \in \mathcal{Q}, \mu \in V}
    \mathbb{E}_{\bu \sim q(\bu, \bz)}
    \left[- \log p(\by \mid \bu) \right] 
    + \KL(q(\bz) \| p(\bz)) \\
    &\mbox{subject to} \quad q(\| \mathbf{r} \|) = \delta(\| \mathbf{r} \|),
\end{empheq}
\end{subequations}
Or equivalently

\begin{subequations}
\begin{empheq}{alignat=2}
    &\argmin_{q(\bz) \in \mathcal{Q}, \mu \in V}
    \mathbb{E}_{\bu \sim q(\bu, \bz)}
    \left[- \log p(\by \mid \bu) \right] 
    + \KL(q(\bz) \| p(\bz)) 
    \label{eq:objective_function}
    \\
    &\mbox{subject to } r(\mathbf{z}_i)=0, \text{ for all finite samples
    $\{\mathbf{z}\}_{i=1}^{M}$ from $q(\bz)$}
\end{empheq}
\end{subequations}

with $r(\bz)$ the scalar function defined by $r(\bz) = \left\|
\mathbf{L}[\bz]\mu(\bz) - \mathbf{f} \right\|_{\mathbf{A}}$.

\subsection{Interpretation as a VAE}
\label{supp:sec:vae}
In this section we provide some further details on how our
our model is to be interpreted as a variational auto-encoder
\citep{kingma, rezende14} with the deocder network modulated by
a supervising PDE problem, in particular we provide details
on two of the simpler aspects not explored in the main paper,
namely the role of the encoder network, and also how the
optimisation framework in Sec. \ref{sec:struct_preserving_VI}
modifies the usual presentation of the VAE.

First we recall that the typical variational problem for a
variational auto encoder has an objective
function of the form
\begin{align*}
    -\mathbb{E}_{\bz \sim q(\bz)}\left[\log p(\by \mid \bz)\right]
    + \KL(q(\bz \mid \by) \| p(\bz) )
\end{align*}
where $p(\by \mid \bz)$ is some high-capacity model for the
conditional probability, typically this will be parameterised
by a neural-network model. We also notice at this point that
our work has had relatively little to say about the 
\emph{encoder} network $q(\bz \mid \by)$, this is an instance
of our desire to allow as much as possible our approach to be used
as a plug-in enhancement to any existing encoder/decoder architecture,
and indeed any of the approaches used in the many body can be
extended to this more general framework by substituting instances
of $q(\bz)$ for $q(\bz \mid \by)$. While we have not examined
this particular aspect of the model, it may still be an interesting area for
future study.

Returning to the specification of the encoder likelihood function
we shall consider the following model for the likelihood of sensor
observations $y_n$ at a spatial coordinate $\mathbf{x}_n$

$$
p(y_n \mid u(\bx_n), \bz) = \mathcal{N}(y_n \mid u(\bx_n), \sigma^2(\bz))
$$

that is we centre the observation distribution on the solution of the
PDE model, but allow for heteroscedastic variance parameterised by
the latent variable $\bz$. 

Then we can write

\begin{align}
    &\mathbb{E}_{\bu \sim q_{\epsilon}(\bu \mid \bz)}
    \left[
    \log \mathcal{N}(y_n \mid u(\bx_n), \sigma^2(\bz))
    \right] \notag \\
    &\qquad = \mathbb{E}_{\bu \sim q_{\epsilon}(\bu \mid \bz)}
    \left[
    -\frac{1}{2\sigma^2(\bz)}(y_n - u(\bx_n))^2 - \frac{1}{2}\log 2\pi \sigma^2(\bz)
    \right] \notag \\
    &\qquad = -\frac{1}{2\sigma^2(\bz)}(y_n - \mu(\bz))^2 
    - \frac{1}{2}\log 2\pi\sigma^2(\bz)
    - \frac{\mathrm{Var}_{q_{\epsilon}}(\bu)}{2 \sigma^2(\bz)} \notag \\
    &\qquad = \log 
    \mathcal{N}(y_n \mid \mu(\bz), \sigma^2(\bz)) + \mathcal{O}(\epsilon^2).
\end{align}

where we have used the fact that the variance term of $q_{\epsilon}$ in
\eqref{eq:approximating_var_cond} is scaled by $\epsilon^2$.

We can now use this to rewrite \eqref{supp:eq:Feps} as

\begin{align}
    &\mathbb{E}_{\bz \sim q(\bz)}
    \left[- \log \mathcal{N}(\by \mid \mu(\bz), \sigma^2(\bz)) \right] 
    + \KL(q(\bz) \| p(\bz)) \notag \\
    &\qquad + \frac{1}{2\epsilon^2 M} 
    \sum_{i=1}^M
    \| \mathbf{L}[\bz_i]\mu(\bz_i) - \mathbf{f} \|_{\mathbf{A}}^2
    + \mathcal{O}(\epsilon^2)
\end{align}

after taking the limit $\epsilon \rightarrow 0$ the final term
will disappear. If we also parameterise $q(\bz)$ as $q(\bz \mid \by)$
then we arrive at the optimisation problem

\begin{subequations}
\begin{empheq}{alignat=2}
    &\argmin_{q(\bz\mid \by) \in \mathcal{Q}, \mu \in V}
    \mathbb{E}_{\bz \sim q(\bz \mid \by)}
    \left[- \log \mathcal{N}(\by \mid \mu(\bz), \sigma^2(\bz)) \right] 
    + \KL(q(\bz \mid \by) \| p(\bz)) 
    \label{supp:eq:objective_function}
    \\
    &\mbox{subject to } r(\mathbf{z}_i)=0, \text{ for all finite samples
    $\{\mathbf{z}\}_{i=1}^{M}$ from $q(\bz \mid \by)$}
\end{empheq}
\end{subequations}

So that \eqref{supp:eq:objective_function} now takes the form of
a mechanistically constrained VAE problem, this is only slightly
more complex than the form \eqref{eq:new_cvi_problem} used in the main body of the
paper and justifies our omission of a fuller discussion of the likelihood
model in the main body of the work. Importantly the variance term
$\sigma^2(\bz)$ is detached from the mean function $\mu(\bz)$,
and it is only the function $\mu(\bz)$ that enters the constraint term
through the constraint term $r(\bz_i)$ which is a function of
the elements $\{ \mu(\bz_i), \mathbf{L}[\bz_i]) \}$.

\section{Mini-patching}
\begin{wrapfigure}{l}{0.45\textwidth}
  \begin{center}
    \includegraphics[width=5cm]{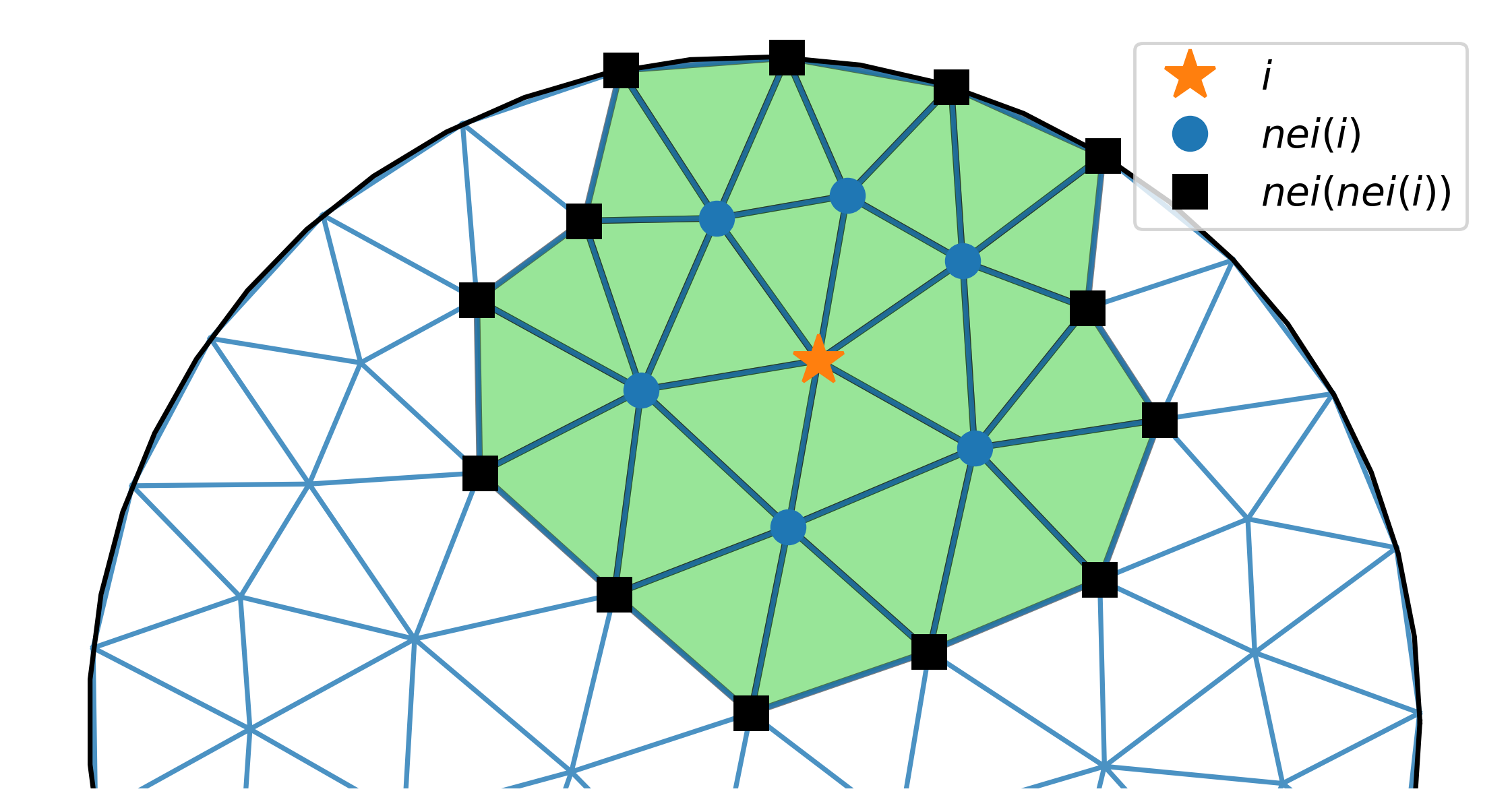}
  \end{center}
  \caption{The mini-patch formed by including all
  elements with a vertex which is in 
  $\mathrm{nei}_{\Delta}(i)$, for some initial vertex.
  This is the reduced mesh over which we evaluate the
  weak-form whenever calculating the component
  $(\mathbf{L}[\bz]\mu(\bz) )_i$ of the discretised weak-form.
  } \label{supp:fig:submesh}
\end{wrapfigure}
In this section we provide some additional details on how
our mini-patching approximation may be used to accelerate
the training of the mechanistically regulated decoder. 
First we recall from Section \ref{sec:fwd_model} that the regularisation term we are seeking to approximate is given by
\begin{align}
    \mathcal{R}(\mathbf{z}) = \| \mathbf{L}[\bz]\mu(\bz) - \mathbf{f} \|_{\bA}^2
    \label{supp:eq:full_reg}
\end{align}
We also recall that the sparsity of the nodal
basis functions translates into sparsity of
the operator $\mathbf{L}[\bz]$, that is we have
\begin{align}
    (\mathbf{L}[\bz]\mu(\bz))_i &= \sum_{j=1}^{\Nnodes} (\mathbf{L}[\bz])_{ij} 
    \mu_j(\bz) \notag \\
    &= \sum_{j \in \mathrm{nei}_{\Delta}(i)} (\mathbf{L}[\bz])_{ij} \mu_j(\bz). 
\end{align}
Noteably to evaluate this term we require only the values
of $(\mathbf{L)}_{ij}$ on the direct neighbours of $i$, in turn this requires performing the quadrature over all mesh elements that have an element of $\mathrm{nei}_{\Delta}(i)$ as a vertex, this relationship
is displayed in Fig. \ref{supp:fig:submesh}. 
Unfortunately, and as mentioned in the main paper, this
does not translate into a useful sparsity pattern for
the regulariser. Indeed we want to evaluate the value
$\mathbf{r}_{\bA}^{(i)}$ where 
\begin{align}
    r^{(i)}_{\bA} = (\WeakForm)_i \sum_{j: (\bA^{-1})_{ij} \neq 0} (\bA^{-1})_{ij}
    (\WeakForm)_j
\end{align}
but in general $(\bA^{-1})$ is a dense matrix, 
that is 
\begin{align}
    \{ j\;:\; (\bA^{-1})_{ij} \neq 0\} = \{ j \; : \; j=1,\ldots, \Nnodes\}
\end{align}
and is therefore of no computational benefit.
However, we nevertheless should expect that
the elements of the covariance matrix 
$\bA^{-1}$ should decrease to zero as
the distance between points increases. Based on 
this idea we propose to use the idea of covariance 
tapering to replace the covariance matrix with a 
localised version, $\Gamma_{\rho}$. 
Infact we chose to make this a hard threshold 
tapering and so define a tapering radius
$\rho > 0$ and define our new covariance term by 
\begin{align}
    (\Gamma_{\rho})_{ij} = \begin{cases}
    (\bA^{-1})_{ij} & \text{if } |\bar{\bx}_i - \bar{\bx}_j| < \rho \\
    0 & \text{otherwise}.
    \end{cases}
\end{align}
The method now proceeds by choosing a vertex $i_p$ uniformly, assembling the matrix $\mathbf{L}[\bz]$ only 
on those values in the implied mini-patch and then
evaluating $r^{(i)}_{\Gamma_{\rho}}$. In principle
$\Gamma$ is still an $\Nnodes \times \Nnodes$ matrix, 
however also note that
because evaluating $r^{(i_p)}$ only requires the non-zero
elements in a row $(\Gamma)_{i_p, :}$ and this is only
those elements which are less that $\rho$ units from
the sampled vertex $i_p$ we can drastically reduce the
computational time by choosing $\rho$ small. This cost
is also a fixed cost that can be done offline before
training and then using a look-up table to index
the vertex $i_p$ and its sparse rows. The 
reduction in time taken to implement our PDE
regularisation method using this tapering has
already been demonstrated 
and discussed in the
transport experiment in Sec.\ref{sec:experiments}, it remains to 
demonstrate that this is still an 
accurate approximation to the original error, and we now 
consider this aspect.

\subsection{Tapering error estimates}
\label{supp:sec:tap_exper}
In Section \ref{sec:experiments}, and in particular in Fig.
\ref{fig:accuracy_exper} we have already demonstrated that
our mini-patching approximation leads to substantially reduced
computation time on the BIP relative to the benchmark HMC
method, and does so while maintaining accuracy. In this
experiment we complement this by
showing that the approximation introduced in 
Sec. \ref{sec:fwd_model}, which allowed us to achieve the
computational benefits, introduces negligible loss of
accuracy compared to the true error $\mathcal{R}(\bz)$.
To carry out the experiment we use the transport equation
\eqref{eq:transport} with constant unit diffusion, spatially 
homogeneous transport vector $\boldsymbol{\tau}(\bx) = (1, 1)^{\top}$
and source $f(\bx) = 1$. We then perturb the model my simulating
from the Gaussian perturbation we introduced in Section 
\ref{sec:relaxing_the_vi_prob}, with scale $\epsilon \in \{0.1, 0.01\}$,
that is we generate samples from the distribution
\begin{align*}
    \mathbf{w} \sim \mathcal{N}(\mathbf{w} \mid \mathbf{0},
    \epsilon^2 \mathbf{A}),
\end{align*}
and form the process $w(\bx) = \sum_{i=1}^{\Nnodes}
(\mathbf{w})_i \phi_i(\bx) $. The perturbed solution is then given by
solving 
\begin{align}
    -\Delta u_{\epsilon}(\bx) 
    + \int_{\Omega} (1, 1)^{\top} \cdot \nabla u_{\epsilon}(\bx) 
    = 1 + w(\bx),
\end{align}
and we denote the coefficients parameterising the solution of this perturbation by $\mathbf{u}_{\epsilon}$. At the same time we 
shall assemble the discretised weak-form corresponding to 
the unperturbed problem, that is we assemble the 
matrix $\mathbf{L}$ and the vector $\mathbf{b}$ with 
elements
\begin{align}
    (\mathbf{L})_{ij} = \int_{\Omega} \nabla \phi_j(\bx) \nabla \phi_i(\bx) dx
    + \int_{\Omega} (1, 1)^{\top} \cdot \nabla\phi_j(\bx) \phi_i(\bx) dx, \qquad 
    (\mathbf{b}_i) = \int_{\Omega} \varphi_i (x) dx,
\end{align}
and then evaluate the error terms
\begin{align}
    \mathbf{r}_{\epsilon, \bA} = 
    \| \mathbf{L} \bu_{\epsilon} - \mathbf{b} \|_{\bA},
    \qquad 
    \mathbf{r}_{\epsilon, \Gamma_{\rho}} =
    \frac{1}{P}\sum_{p=1}^{P}
    \| \mathbf{L} \bu_{\epsilon} - \mathbf{b} \|_{(\Gamma_{\rho})_{{i_p},:}}, \label{supp:eq:taper_targets}
\end{align}
where we use the notation 
$\| \mathbf{w} \|_{(\Gamma_{\rho})_{{i_p},:}}$ to denote
the product of the two scalars
\begin{align*}
    (\mathbf{w})_{i_p} \cdot \langle (\Gamma_{\rho})_{{i_p},:},
    \mathbf{w} \rangle_{\mathbb{R}^{\Nnodes}}.
\end{align*}
The first term in \eqref{supp:eq:taper_targets} is 
our target error, and the second of these
is our tapered mini-patch approximation to this error. 
By perturbing we make sure this first term is not 
trivially equal to zero, and so investigate the behaviour of 
our approximation around the target value. In this experiment 
we report the absolute difference of these two estimates, that is
we report the absolute error 
$|\mathbf{r}_{\epsilon, \bA} 
- \mathbf{r}_{\epsilon, \Gamma_{\rho}}|$ obtained from
a total of $10$ different samples of the perturbation
process $w(\bx)$.

\begin{figure}[h]
    \centering
    \begin{subfigure}[b]{0.45\linewidth}
        \centering
        \includegraphics[width=\textwidth]{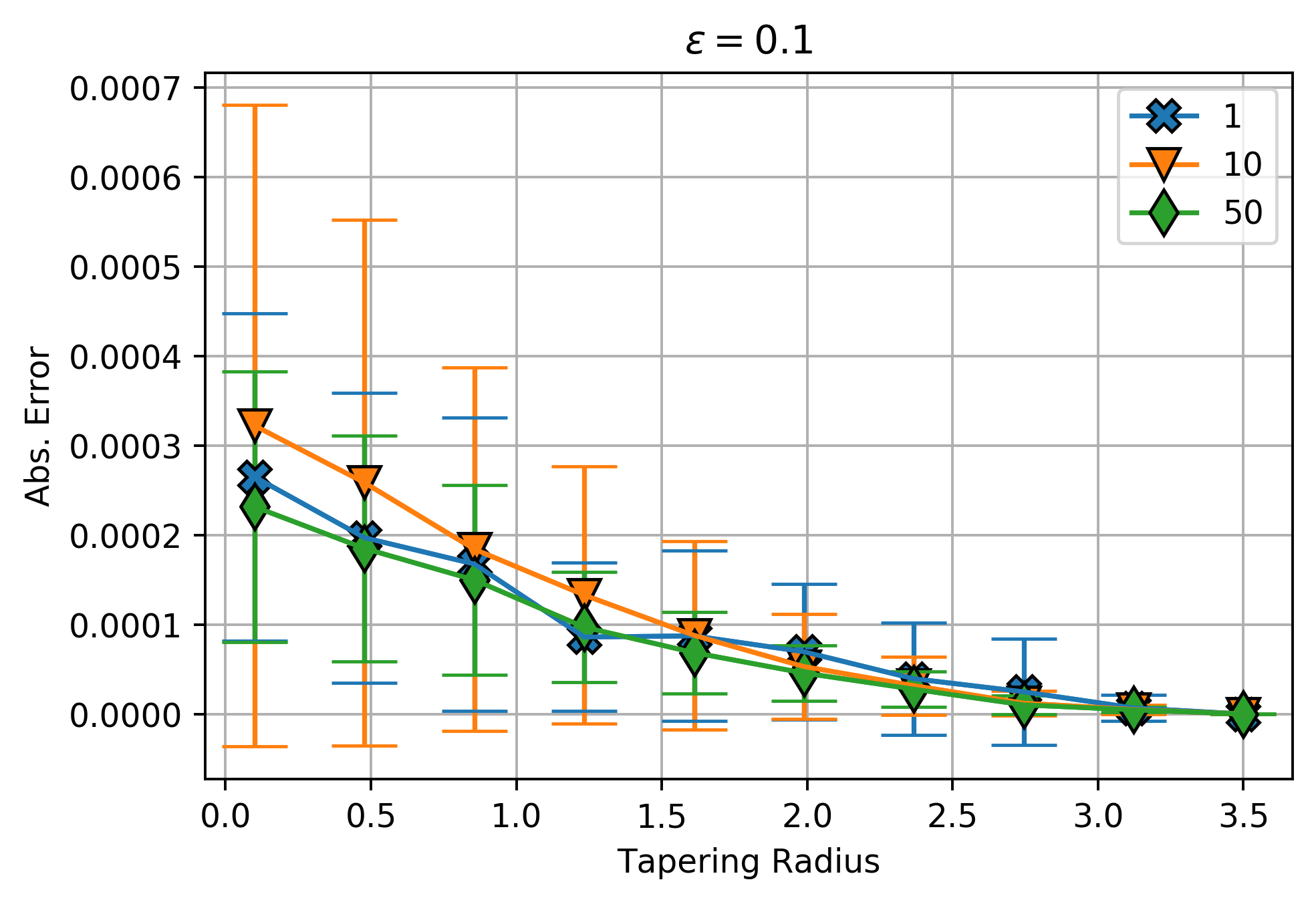}
        \caption{Approximation error, $\epsilon=0.1$}
        \label{supp:fig:mp_exper_a}
    \end{subfigure}
    \hfill
    \begin{subfigure}[b]{0.45\linewidth}
        \centering
        \includegraphics[width=\textwidth]{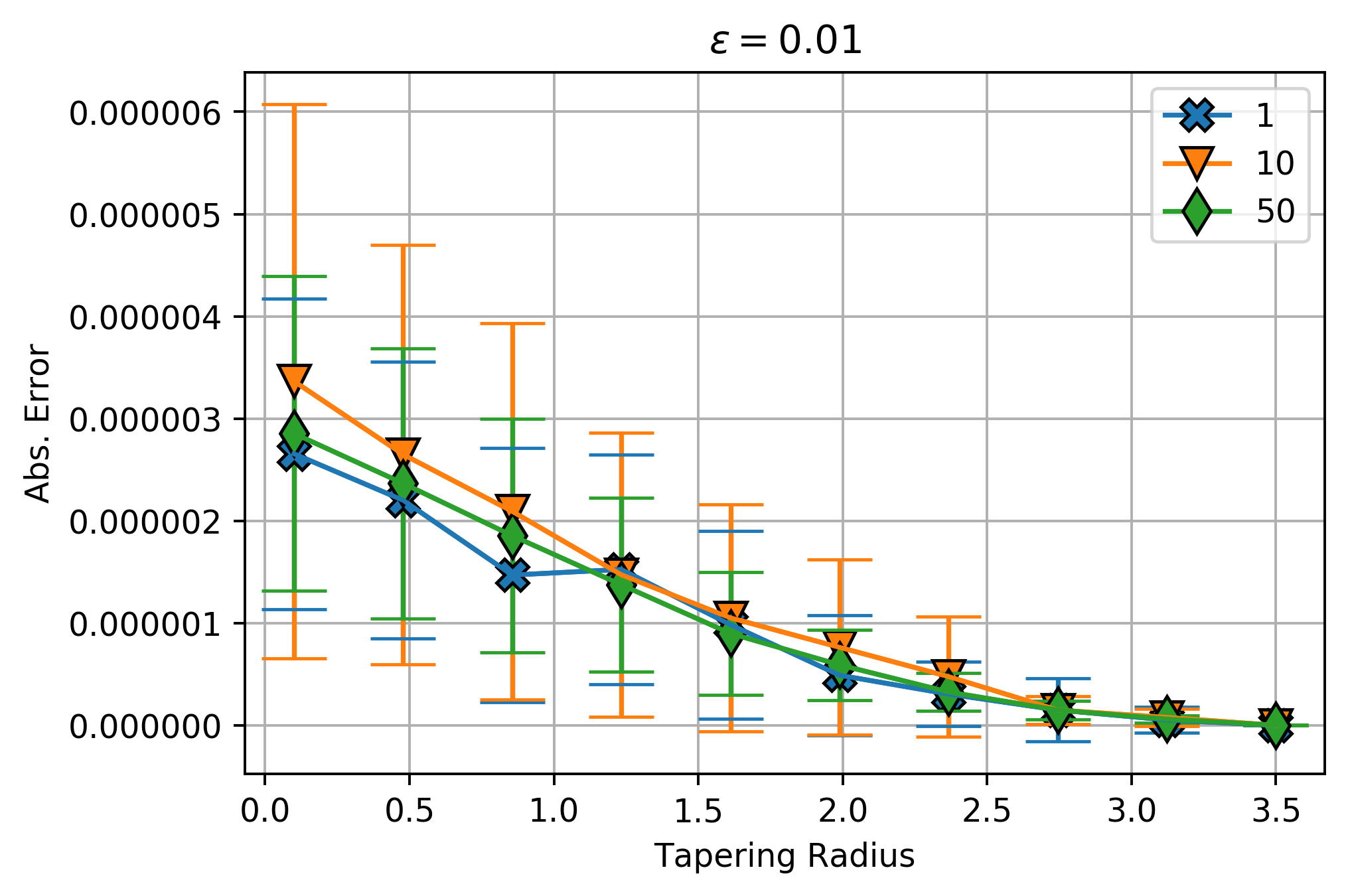}
        \caption{Approximation error, $\epsilon=0.01$}
        \label{supp:fig:mp_exper_b}
        \vspace*{2mm}
    \end{subfigure}
    \begin{subfigure}[b]{0.45\linewidth}
        \centering
        \includegraphics[width=\textwidth]{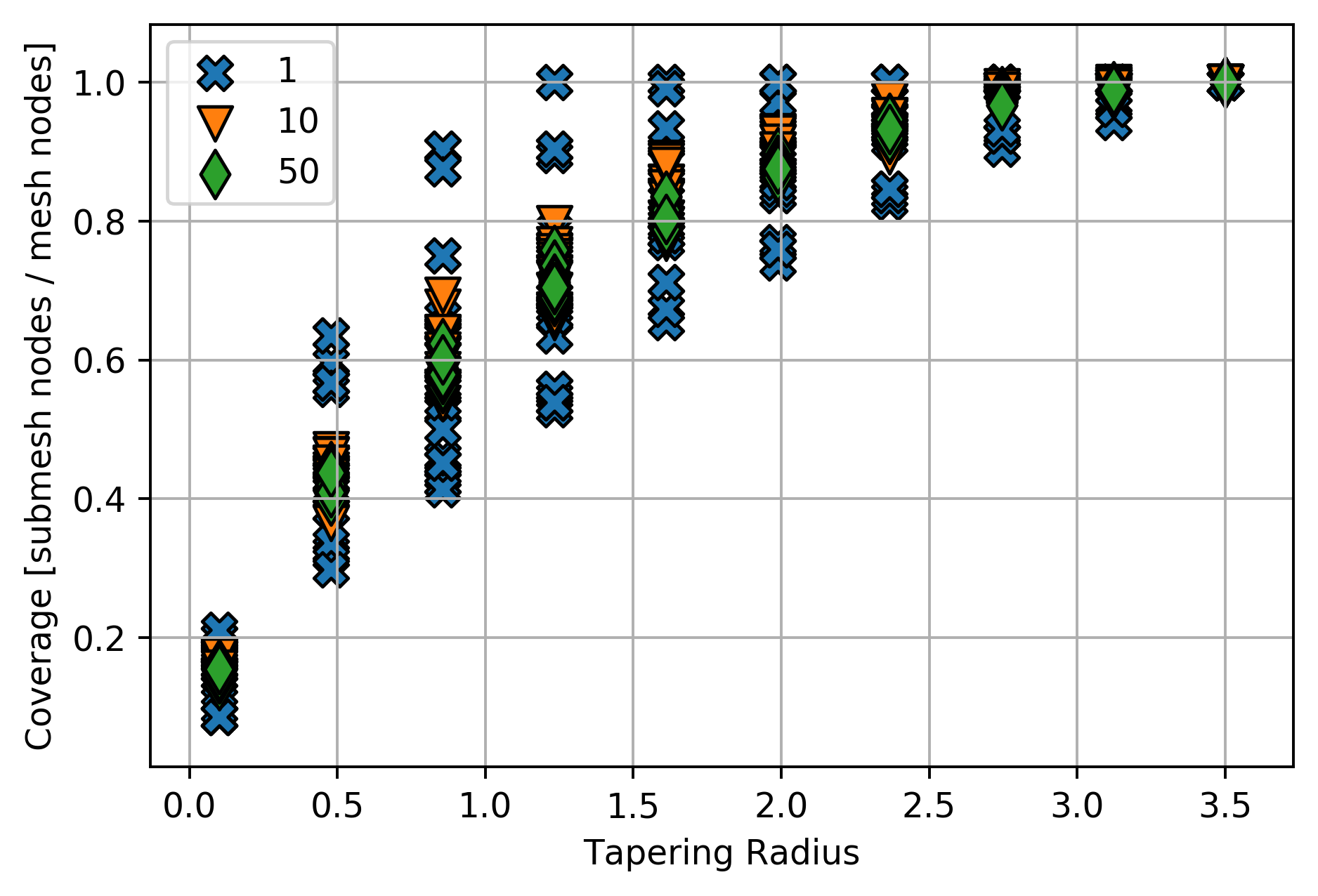}
        \caption{Average mesh coverage}
        \label{supp:fig:rad_vs_cov}
    \end{subfigure}
    \caption{Absolute error of the tapered mini-patching approximation as a
    function of the tapering radius, and the number of meshes sampled, $P
    \in \{1, 10, 50\}$}
    \label{supp:fig:mp_exper}
\end{figure}

The results are displayed in Fig. \ref{supp:fig:mp_exper}, 
where we can observe that as one would expect the absolute 
error of the tapered
approximation decreases to zero as the tapering radius increases since
the tapered mesh converges to the full mesh. Importantly for the
application of our method we note two important details; (i) the
absolute error is small even for the smallest sub-meshes, and indeed
is orders or magnitude lower than the perturbation error $\epsilon$, 
and (ii) that even the single sample estimate is within the error
limit of the methods using more samples. Given this second remark we 
choose to use a single mesh sample in all of the remaining 
experiments reported in this paper. In Fig. \ref{supp:fig:rad_vs_cov} we 
also plot the tapering radius versus the average fraction of the whole 
mesh covered by a mini-patch, demonstrating that the chosen radius 
ranges plotted in Fig. \ref{supp:fig:mp_exper} were sufficient to cover 
the full spectrum of relevant mesh portions.

\section{Further details of the experiments}
In this section we provide some additional details for the
experiments reported in Sec. \ref{sec:experiments} of the
main paper, including the specific architecture used
in each instance. All models were implemented in
Tensorflow \citep{tensorflow2015-whitepaper} including
a Tensorflow implementation of the FEM, 
full code for which is available from the authors' 
website. Optimisation was done using the Adam 
optimiser with default parameters.

\subsection{Jura experiment}
\subsubsection*{Model specification} 
The form of the PDE mechanisms used to supervise the experiments
was given in Fig. \ref{fig:vae} but for convenience we restate them
here
\begin{figure}[h]
    \centering
    \begin{subfigure}[b]{.45\linewidth}
        \centering
        \begin{align}
        \mathcal{L}u = -\nabla \cdot (a(\bx) \nabla u) 
                + \boldsymbol{\tau}^{\top} \cdot \nabla u
        \end{align}
        \caption{Mechanistic form of \texttt{VAE-DT}}
    \end{subfigure}
    \hfill
    \begin{subfigure}[b]{.45\linewidth}
        \centering
        \begin{align}
        \mathcal{L}u = -\nabla \cdot (a(\bx, u, \nabla u) \nabla u)
        \end{align}
        \caption{Mechanistic form of \texttt{VAE-NLD}}
    \end{subfigure}
    \caption{Mechanistic operators used to regularise
    the base VAE model in the Jura experiment.}
\end{figure}

\subsubsection*{Inference}
As emphasised in
the main paper our method is intended to augment a 
standard model, therefore all of the VAE variants used
the same encoder/decoder network which take the forms given
in Fig. \ref{fig:vae_encoder_decoder_forms} the models are
then trained by adding our pde regularised loss to the
standard variational loss as described in Sec. 
\ref{sec:relaxing_the_vi_prob} of the paper.

Our objective is to use the presence of the secondary metals,
to predict the concentration of a primary, that is we are
attempting to learn a variational approximation to the 
conditional distribution $p(y^{(p)}(\bx) \mid \bx, 
\mathbf{s}^{(p)}(\bx))$ where $p$ is a primary metal in 
$\{\text{CD}, \text{CU}, \text{PB}, \text{CO} \}$, and
$\mathbf{s}^{(p)}(\bx)$ is the value of a collection of
secondary metals at that same spatial coordinate, a table
of the primary and second metals from \citep{alvarez_2009}
is given in \ref{supp:tab:metals}. 

\begin{table}[ht]
\caption{Primary and secondary metals for the 
Jura experiment \citep{goovaerts, alvarez_2009}}
\label{supp:tab:metals}
\vskip 0.15in
\begin{center}
\begin{small}
\begin{sc}
\begin{tabular}{ll}
\toprule
Primary & Secondary \\
\midrule 
Cd & Pb, Ni, Zn \\
Cu & Ni, Zn \\
Pb & Cu, Ni, Zn \\
Co & Ni, Co \\
\bottomrule
\end{tabular}
\end{sc}
\end{small}
\end{center}
\vskip -0.1in
\end{table}

The conditional distribution $p(y^{(p)}(\bx) \mid \bx, 
\mathbf{s}^{(p)}(\bx))$ is therefore the target of
our decoder $q(\by \mid \bz, \bx, \mathbf{s})$, and we learn
this factor for a base VAE model, and the PDE regularising
operators in Fig. \ref{fig:vae} using the objective
function presented in Sec. \ref{sec:struct_preserving_VI},
see also Sec. \ref{supp:sec:vae}.

\begin{figure}
\begin{subfigure}[b]{\linewidth}
\centering
\begin{verbatim}
Model: "VAE encoder"
_________________________________________________________________
Layer (type)                 Output Shape              Param #   
=================================================================
input_10 (InputLayer)        [(None, 5)]               0         
_________________________________________________________________
dense_1 (Dense)              (None, 128)               768       
_________________________________________________________________
dense_2 (Dense)              (None, 128)               16512     
_________________________________________________________________
dense_3 (Dense)              (None, 128)               16512     
_________________________________________________________________
dense_4 (Dense)              (None, 65)                8385      
_________________________________________________________________
qz (MultivariateNormalTriL)  ((None, 10), (None, 10))  0         
=================================================================
Total params: 42,177
Trainable params: 42,177
Non-trainable params: 0
\end{verbatim}
\caption{Encoder network used in the VAE for the 
Jura experiment}
\label{fig:vae_encoder}
\end{subfigure}
\vspace{10mm}
\begin{subfigure}[b]{\linewidth}
\centering
\begin{verbatim}

Model: "VAE decoder"
_________________________________________________________________
Layer (type)                 Output Shape              Param #   
=================================================================
input_18 (InputLayer)        [(None, 12)]              0         
_________________________________________________________________
dense_1 (Dense)              (None, 8)                 104       
_________________________________________________________________
dense_2 (Dense)              (None, 16)                144       
_________________________________________________________________
dense_3 (Dense)              (None, 32)                544       
_________________________________________________________________
dense_4 (Dense)              (None, 16)                528       
_________________________________________________________________
dense_5 (Dense)              (None, 8)                 136       
_________________________________________________________________
param_obs_dist (Dense)       (None, 2)                 18        
_________________________________________________________________
obs_dist (DistributionLambda ((1, None), (1, None))    0         
=================================================================
Total params: 1,474
Trainable params: 1,474
Non-trainable params: 0
\end{verbatim}
\caption{Decoder network for the VAE model}
\end{subfigure}
\caption{Encoder and decoder networks used for the VAE
applied to the Jura dataset. All Dense networks use
ReLU activations apart from the final ones in each
sequential model which use linear activations. (a) 
The encoder network returns a 
\texttt{MultivariateNormalTriL} object corresponding to 
the variational factor $q(\bz \mid \by)$. (b) The
decoder network returns a collection of independent
normal observation models $p(y(\bx_n) \mid 
\bz, \bx_n) = \mathcal{N}(y(\bx_n) \mid 
\mu(\bx, \bz), \sigma^2(\bx, \bz))$ where $\mu(\bx, \bz)$
is our PDE regularised forward surrogate, and obtained by
a slice \texttt{[..., :1]} into the output of
the \texttt{param\_obs\_dist} dense network. In this
instance there are 3 secondary metals plus the spatial
coordinate so the input to the encoder is shape 
5.}
\label{fig:vae_encoder_decoder_forms}
\end{figure}

\subsection{Aquifer experiment}
\begin{figure}
    \centering
    \begin{subfigure}[b]{.46\linewidth}
        \centering
        \begin{subequations}
            \begin{align}
                -\nabla \cdot (a(\bx) \nabla u) &= f(\bx) 
                \label{supp:eq:linpde1}\\
                \log a(\bx) &\sim \mathcal{GP}(0, k(\bx, \bx')) \\
                f(\bx) &\sim \mathcal{GP}(0, k_{f}(\bx, \bx'))
            \end{align}
        \end{subequations}
        \caption{Mechanistic form of \texttt{GP-D}}
    \end{subfigure}
    \hfill
    \begin{subfigure}[b]{.46\linewidth}
        \begin{subequations}
            \begin{align}
                -\nabla \cdot (a(\bx) \nabla u)  \notag \\
                + \boldsymbol{\tau}^{\top} \cdot \nabla u
                &= f(\bx) \label{supp:eq:linpde2}\\
                \log a(\bx) &\sim \mathcal{GP}(0, k(\bx, \bx')) \\
                f(\bx) &\sim \mathcal{GP}(0, k_{f}(\bx, \bx')) \\
                \boldsymbol{\tau} &\sim \mathcal{N}(\mathbf{0}, \mathbf{I})
            \end{align}
        \end{subequations}
        \caption{Mechanistic form of \texttt{GP-DT}}
    \end{subfigure}
    \caption{Strict mechanistic versions of the 
    models used in the aquifer experiment. Conditional
    on the diffusion coefficient $\mathbf{a}(\bx)$, both
    \eqref{supp:eq:linpde1} and \eqref{supp:eq:linpde2}
    are linear PDEs forced by Gaussian noise, and so these
    models are conditionally Gaussian processes. To 
    improve computational efficiency, and to allow for
    the possibility of model misspecification we replace
    the implicit solution $u(\bx)$ with our forward
    surrogate. In the experiments $k$ and $k_f$ are both
    taken to be Matern 5/2 kernels}
    \label{fig:aquifer_mechanisms}
\end{figure}

\subsubsection*{Model specification} 
We define two models \texttt{GP-D} and \texttt{GP-DT} to
carry out mechanistically informed modelling of the
aquifer. The mechanistic structure of each of these models 
is presented in Fig. \ref{fig:aquifer_mechanisms}. 
Conditionally both of 
these models are linear PDEs, so that
we can quite naturally consider these models as being
mechanistically structured hierarchical GP models, or that is
to say deep Gaussian processes \citep{damianou_2013}, for this
reason we continue to refer to these model as
\texttt{GP-D} and \texttt{GP-DT} respectively once we 
replace the true forward map with the GP surrogate. 
The exact model in both instances is replaced with a forward surrogate 
which has the form given in Fig. \ref{fig:aquifier_fwd_surrogate}.
During training we starting with setting $\epsilon = 0.1$ and
gradually decrease it to $\epsilon=0.01$ over a period of 
1000 epochs, this is repeated until the overall optimisation
is terminated. By restarting the constraint in this manner we
prevent the method from concentrating on trivial solutions to the
PDE problem.

\begin{figure}
\centering
\begin{verbatim}
Model: "fwd_surrogate"
_________________________________________________________________
Layer (type)                 Output Shape              Param #   
=================================================================
well_loc (InputLayer)        [(None, 2)]               0         
_________________________________________________________________
dense_1 (Dense)              (None, 32)                96        
_________________________________________________________________
dense_2 (Dense)              (None, 64)                2112      
_________________________________________________________________
dense_3 (Dense)              (None, 64)                4160      
_________________________________________________________________
dense_4 (Dense)              (None, 1)                 65        
=================================================================
Total params: 6,433
Trainable params: 6,433
Non-trainable params: 0 
\end{verbatim}
\caption{Forward surrogate model for the aquifer experiment, 
all Dense layers use a sigmoid activation apart from the
final layer which has no activation. This model is a function
of the spatial coordinate only and outputs the mean of the
ground water level observation distribution.}
\label{fig:aquifier_fwd_surrogate}
\end{figure}

\subsubsection*{Inference} 
Given the close analogy of the hierarchical mechanistic
models displayed in Fig. \ref{fig:aquifer_mechanisms} we
train this model in a manner similar to that used for
deep GPs using the doubly-stochastic approach \citep{dsdgp}.
That is we replace the GPs $\log a(\bx)$ and $f(\bx)$
with their sparse variational GP (SVGP) approximations
\citep{hensman}, and then free-form optimise the 
parameterised distributions of the inducing 
variables inside of the
variational framework we have introduced in 
Sec. \ref{sec:struct_preserving_VI}, since
$\tau$ is also a Gaussian process, albeit a trivial
one, we also replace this model component with a free-form
Gaussian factor with the same event shape. For the log-diffusion
GP and the source GP we use $50$ inducing points. 
Let $g(\bx) = \log a(\bx)$, then we aim to estimate the 
following variational factor
\begin{align}
        q(\bff, \mathbf{g}, \boldsymbol{\tau})
        &= q(\bff)q(\mathbf{g})q(\boldsymbol{\tau}) \notag \\
        &=\mathcal{N}(\bff \mid \mathbf{m}_f, \mathbf{S}_f)
\mathcal{N}(\mathbf{g} \mid \mathbf{m}_g, \mathbf{S}_g)
\mathcal{N}(\boldsymbol{\tau} \mid \mathbf{m}_\tau, \mathbf{S}_\tau)
\end{align}
where the parameters $\{\mathbf{m}_\theta, 
\mathbf{S}_{\theta} \}$, $\theta \in \{f, g, \tau\}$
are the mean and covariance of the Gaussian variational
factors we optimise for, corresponding to the source,
log-diffusion coefficient, and in the case of
\texttt{GP-DT} the transport vector field, respectively.
In common with the approach taken for DGP models we
do not further allow these learned factors to depend
on the inputs, that is we unlike the previous experiment
we do not include an encoder network and this model is
purely a decoder from the variational learned factor.
Finally the conditional probability of the observations
takes the form $p(y(\bx) \mid \mu) = \mathcal{N}(
y(\bx) \mid \mu(\bx), \sigma^2)$, that is the ground water
level $y(\bx)$ at location $\bx$ is Gaussian distributed with
mean function given by the forward surrogate and spatially
homogeneous variance.

\subsubsection*{Additional Figures} 
\begin{figure}
    \centering
    \begin{subfigure}[b]{0.45\linewidth}
        \centering
        \includegraphics[width=\textwidth]{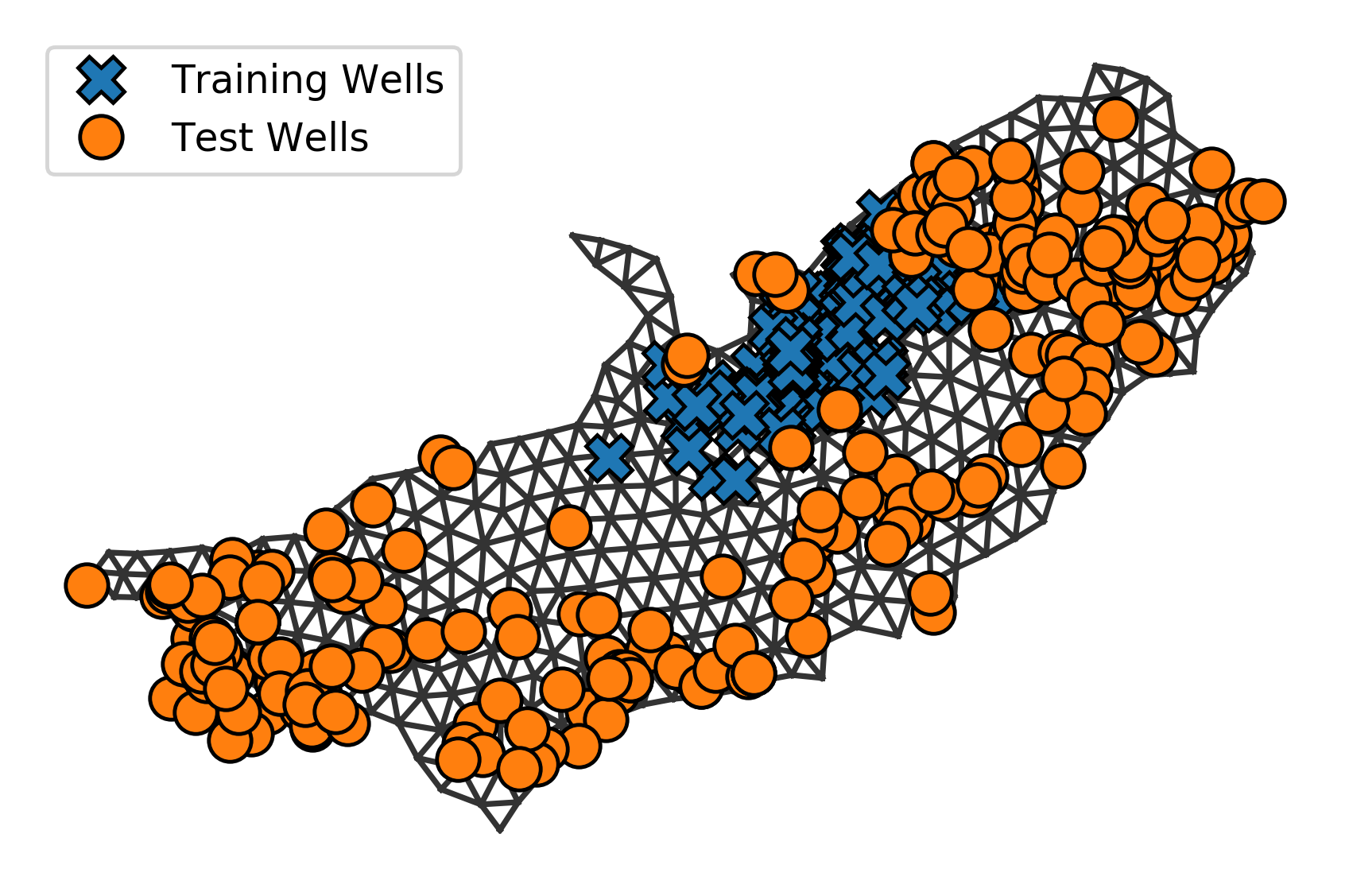}
        \caption{Well locations and regional mesh}
    \end{subfigure}
    \begin{subfigure}[b]{0.45\linewidth}
        \centering
        \includegraphics[width=\textwidth]{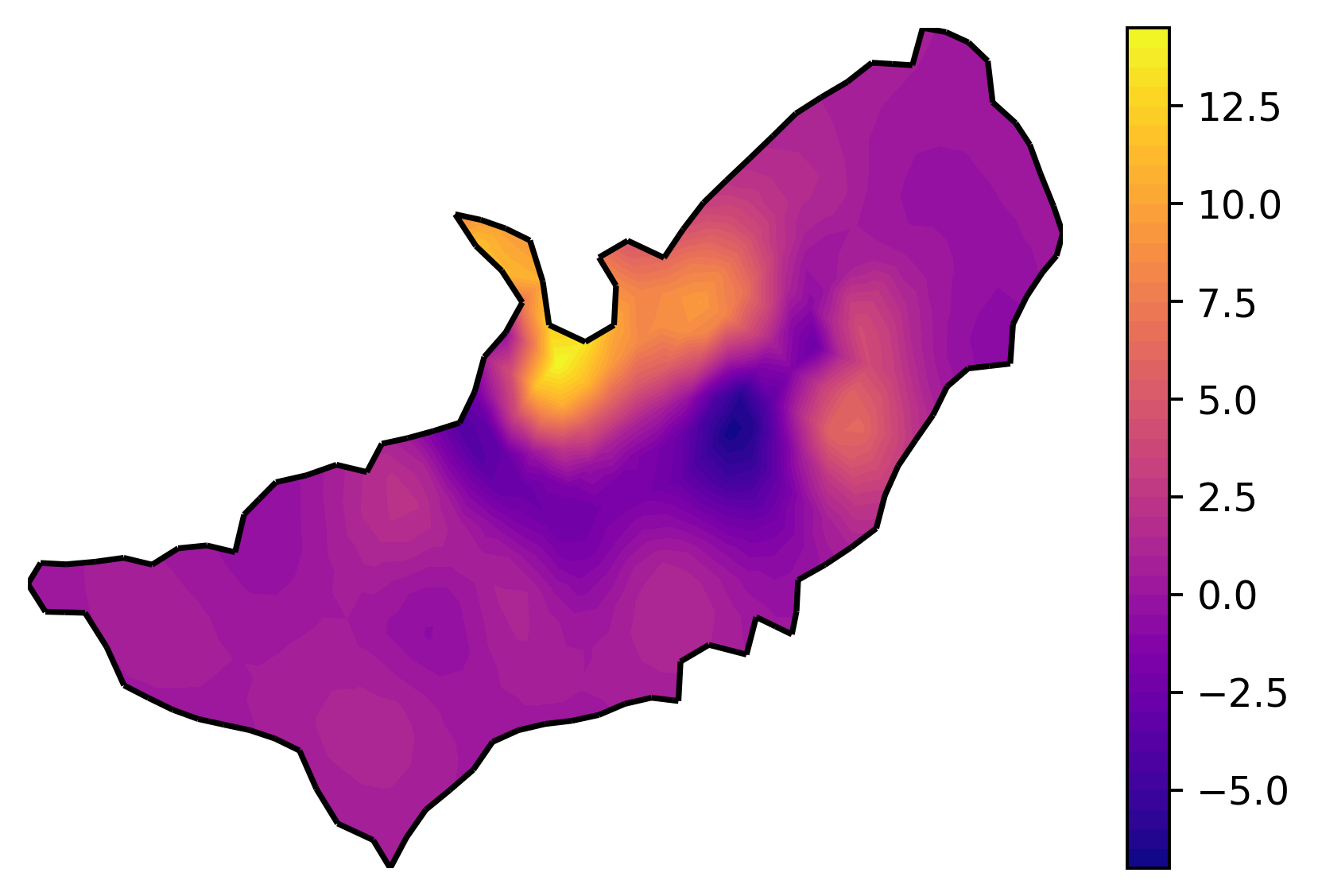}
        \caption{Mean source, $\mathbb{E}_{q(\bff)}[f(\bx)]$}
        \label{fig:aquifer_add_figures_b}
    \end{subfigure}
        \begin{subfigure}[b]{0.45\linewidth}
        \centering
        \includegraphics[width=\textwidth]{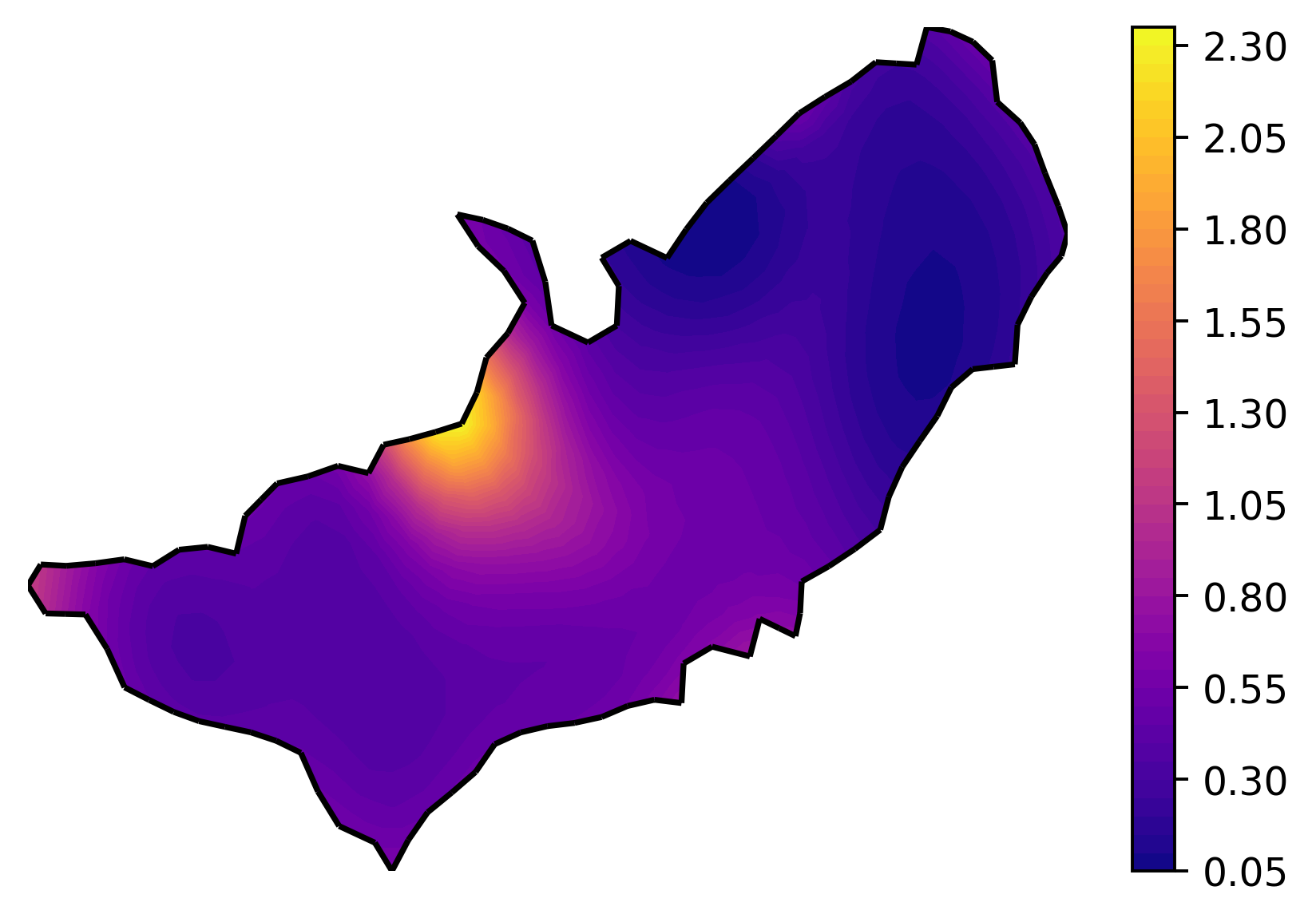}
        \caption{Mean diffusion coefficient, $\mathbb{E}_{q(\mathbf{g})}[e^{g(\bx)}]$}
        \label{fig:aquifer_add_figures_c}
    \end{subfigure}
    \caption{Additional figures for the aquifer experiment.
    (a) displays the locations of the training and test well
    sites. (b) presents the mean of the source
    from our \texttt{GP-DT}, and (c) presents the mean
    diffusion coefficient}
    \label{fig:aquifer_add_figures}
\end{figure}

Additional figures for the
Aquifer experiment are dispalyed in Fig. 
Fig. \ref{fig:aquifer_add_figures}. We would draw particular attention
to the fact that the training wells from \citep{ackerman} cover
only a small region of the local aquifer around the Idaho 
national laboratory. This makes the extrapolation problem to the
test set much more challenging unless the model embodies some
physical structure. A similar insight is contained in the work
of \citep{arno}, but note that we don't require knowledge of
exactly what the process values should be on the boundary, and
so do not need to embed this information as a hard constraint.

\begin{figure}
    \centering
    \begin{subfigure}[b]{0.45\linewidth}
        \centering
        \includegraphics[width=\textwidth]{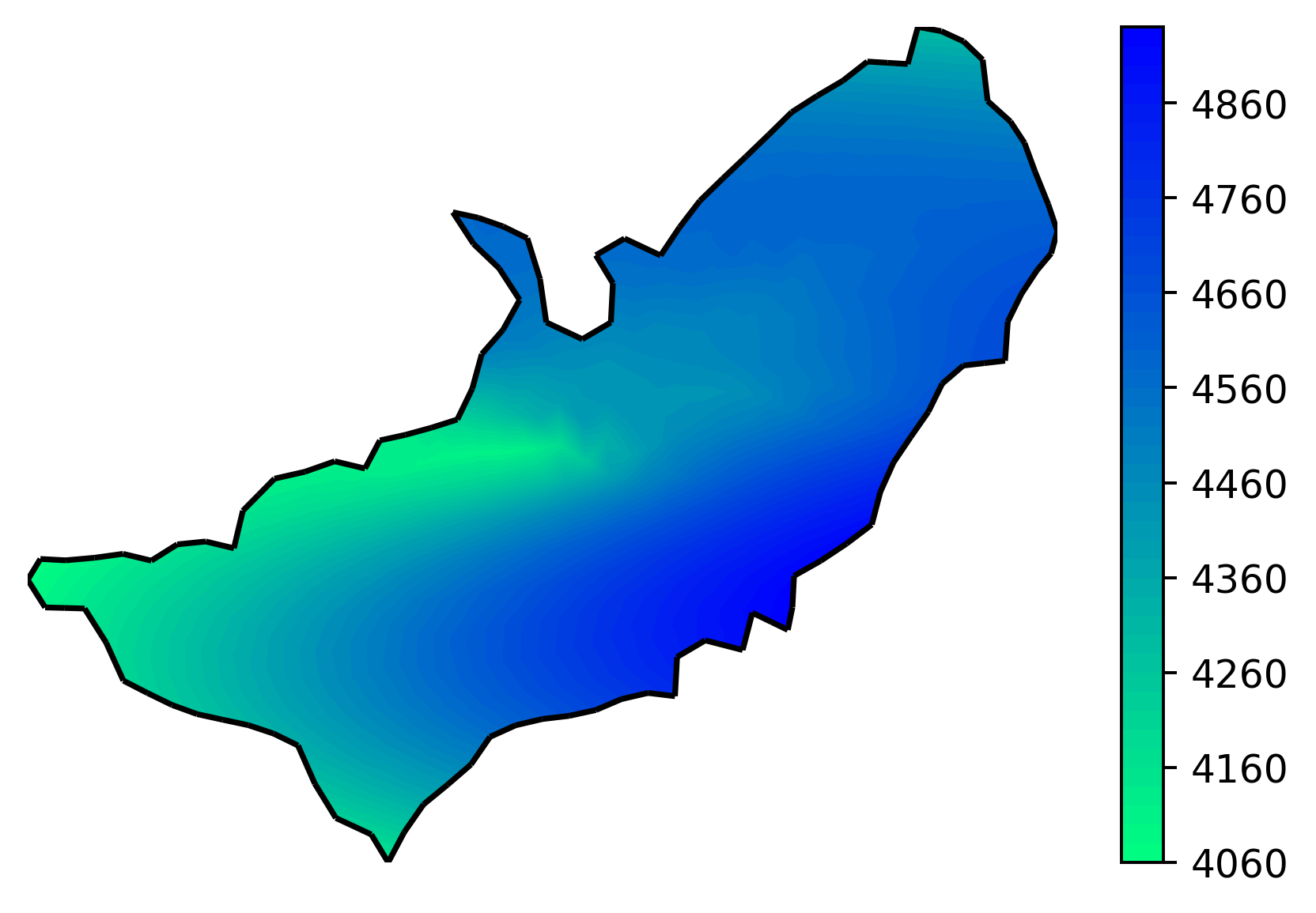}
        \caption{GP predicted ground water level}
        \label{fig:aquifer_pred_a}
    \end{subfigure}
    \begin{subfigure}[b]{0.45\linewidth}
        \centering
        \includegraphics[width=\textwidth]{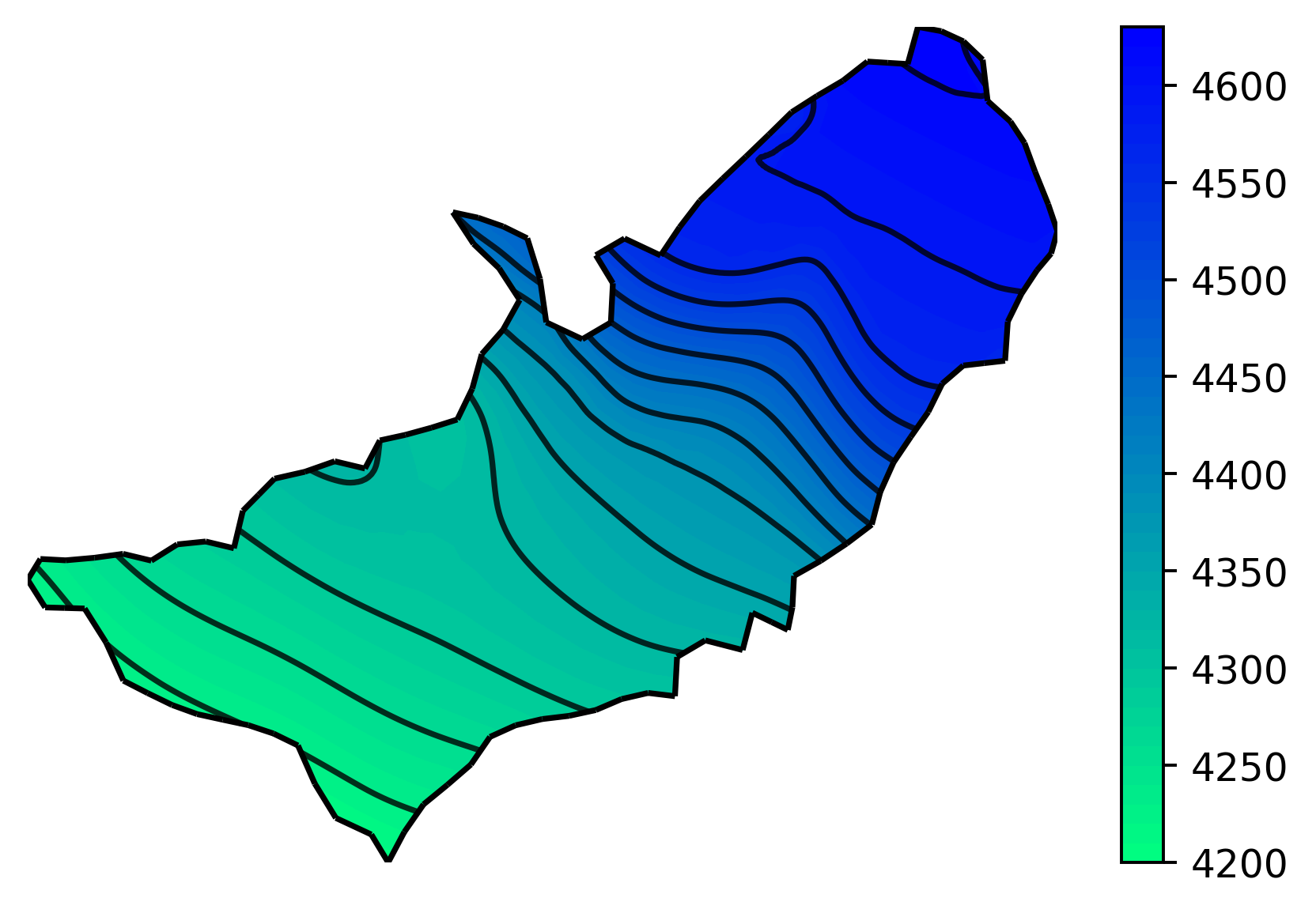}
        \caption{\texttt{GP-DT} predicted ground water level}
        \label{fig:aquifer_pred_b}
    \end{subfigure}
    \caption{Additional ground water prediction 
    figures for the aquifer experiment}
    \label{fig:aquifer_comparison}
\end{figure}

Our physically informed model is able to therefore learn a mechanism
and so carry out sensible extrapolation, but this does not come
at the expense of unrealistic prior restrictions, noteably when
looking at the learned source function Fig. \ref{fig:aquifer_add_figures_b}
we see that away from the data the source estimate is zero, and a 
similar observation can be made from the learned diffusion coefficient 
Fig. \ref{fig:aquifer_add_figures_c}. By fitting a mechanistic model
we learn parameters that fit the training data, avoid unrealistic 
prior assumptions, but crucially lead to region-wide generalisation. 
This is contrasted with the ground water level predictions from the 
mechanism free GP model in Fig. \ref{fig:aquifer_comparison}, 
which over-trains on the well locations and possesses no mechanism 
by which to regularise this behaviour.

\end{document}